\documentclass[10pt,journal,compsoc]{IEEEtran}
\usepackage{amssymb}
\usepackage{graphicx}
\usepackage{epstopdf}
\usepackage{gensymb}
\usepackage{color}
\usepackage{amsmath}
\usepackage{bm}
\usepackage{etoolbox}
\usepackage{cite}
\usepackage{array}
\usepackage{booktabs}
\usepackage{multirow}
\usepackage{comment}
\usepackage{subcaption}
\usepackage[ruled,vlined]{algorithm2e}
\usepackage{mathrsfs}
\usepackage{algpseudocode}
\usepackage{hyperref}
\usepackage{balance}
\hypersetup{
    colorlinks=true,
    linkcolor=blue,
    filecolor=magenta,      
    urlcolor=cyan,
}
 
\alglanguage{pseudocode}

\usepackage[framemethod=TikZ]{mdframed}

\algnewcommand{\Initialize}[1]{%
  \State \textbf{Initialize:}
  \Statex \hspace*{\algorithmicindent}\parbox[t]{.8\linewidth}{\raggedright #1}
}

\usepackage{threeparttable}
\usepackage{mathtools}
\usepackage{hyperref}
\usepackage{pifont}
\usepackage{listings}

\usepackage{url}
\urldef{\mailsa}\path|{alfred.hofmann, ursula.barth, ingrid.haas, frank.holzwarth,|
	\urldef{\mailsb}\path|anna.kramer, leonie.kunz, christine.reiss, nicole.sator,|
	\urldef{\mailsc}\path|erika.siebert-cole, peter.strasser, lncs}@springer.com|    

\usepackage[utf8]{inputenc}
\usepackage[english]{babel}

\usepackage{amsthm}

\usepackage{soul}

\theoremstyle{definition}

\newtheorem{preposition}{Preposition}

\definecolor{R}{RGB}{0,0,150}

\theoremstyle{remark}

\ifCLASSINFOpdf

\begin{document}

\title{Evaluation and Optimization of Distributed Machine Learning Techniques for \\ Internet of Things}

\author{
Yansong Gao, Minki Kim, Chandra Thapa, Alsharif Abuadbba, \\ Zhi Zhang, Seyit Camtepe, Hyoungshick Kim, and Surya Nepal
  
\IEEEcompsocitemizethanks{\IEEEcompsocthanksitem Y.~Gao, C. Thapa, Z. Zhang, and S. Camtepe are with Data61, CSIRO, Sydney, Australia. e-mail: \{garrison.gao;zhi.zhang;chandra.thapa; seyit.camtepe\}@data61.csiro.au.}
\IEEEcompsocitemizethanks{\IEEEcompsocthanksitem M.~Kim, H. Kim are with Department of Computer Science and Engineering, College of Computing, Sungkyunkwan University, South Korea, and Data61, CSIRO, Sydney, Australia and  e-mail:\{mk7777; hyoung\}@skku.edu.}
\IEEEcompsocitemizethanks{\IEEEcompsocthanksitem A. Abuadbba, S.~Nepal are with Data61, CSIRO, Sydney, Australia, and Cyber Security Cooperative Research Centre, Australia. e-mail: \{sharif.abuadbba; surya.nepal\}@data61.csiro.au.}

}
 
\IEEEtitleabstractindextext{		
\begin{abstract}
Federated learning (FL) and split learning (SL) are state-of-the-art distributed machine learning techniques to enable machine learning training without accessing raw data on clients or end devices. However, their \emph{comparative training performance} under real-world resource-restricted Internet of Things (IoT) device settings, e.g., Raspberry Pi, remains barely studied, which, to our knowledge, have not yet been evaluated and compared, rendering inconvenient reference for practitioners. This work firstly provides empirical comparisons of FL and SL in real-world IoT settings regarding (i) learning performance with heterogeneous data distributions and (ii) on-device execution overhead. Our analyses in this work demonstrate that the learning performance of SL is better than FL under an imbalanced data distribution but worse than FL under an extreme non-IID data distribution. 
Recently, FL and SL are combined to form splitfed learning (SFL) to leverage each of their benefits (e.g., parallel training of FL and lightweight on-device computation requirement of SL). 
This work then considers FL, SL, and SFL, and mount them on Raspberry Pi devices to evaluate their performance, including training time, communication overhead, power consumption, and memory usage.
Besides evaluations, we apply two optimizations. Firstly, we generalize SFL by carefully examining the possibility of a hybrid type of model training at the server-side. The generalized SFL merges sequential (dependent) and parallel (independent) processes of model training and is thus beneficial for a system with large-scaled  IoT devices, specifically at the server-side operations. %
Secondly, we propose pragmatic techniques to substantially reduce the communication overhead by up to four times for the SL and (generalized) SFL.

\end{abstract}
}
\maketitle

\section{Introduction}\label{sec:intro}

Attributing to its stunning performance, deep learning (DL) has enabled various applications ranging from image classification, object detection, speech recognition to disease diagnosis, financial fraud detection~\cite{lecun2015deep,wang2017adversary,tang2016deep}. One major factor in achieving high accuracy is usually to exploit big data to learn high-level features. The intuitive means is to gather the data centrally and then perform the DL model training. However, data can often be highly private or sensitive; for example, data collected from medical sensors~\cite{huang2019patient} and microphones~\cite{i2smicrophone} would be such cases. Consequently, users may resist sharing their data with service/cloud providers to build a DL model. In addition, the data aggregator must pay great attention to the data regulations such as the General Data Protection Regulation (GDPR)~\cite{gdpr}, and California Privacy Rights Act (CPRA)~\cite{cpra}. On the other hand, the centralized data could be mishandled or incorrectly managed by service providers---e.g., incidentally accessed by unauthorized parties~\cite{amazonstranger}, or used for unsolicited analytic, or compromised through the network and system security vulnerabilities---resulting in the data breach~\cite{shastri2019seven,biometricbreach}. Therefore, there is a demand for training an ML model without aggregating and accessing sensitive raw data resided on the client-side~\cite{fedlearningMcMahan17,fed2,wu2020safa,gupta2018distributed,nopeeksplitNN}.

To resolve the above problem, distributed machine learning techniques have been developed to train a joint/global model with no direct access to the decentralized locally resided raw data. Such techniques are of great appeal to distributed system applications to reap the benefits from rich data yielded by IoT devices in distributed IoT architectures. Distributed learning techniques keep the data locally and utilize private data (e.g., medical records, voice records, and text inputs) during the learning process to reduce privacy leakage risks. However, there is still a significant gap in evaluating the training performance of those techniques concerning their practicality in the IoT-enabled distributed systems constituted by resource-constrained devices.  

\vspace{-0.3cm}
\subsection{Limitations} As most IoT devices are resource-restricted, the ML training \textit{inference} and \textit{training} overhead should be firstly evaluated and then optimized. Currently, most studies are focusing on the on-device \textit{inference} performance of resource-restricted. In this case, models are in fact trained with resource-rich computing platforms, while training on resource-restricted computing platforms is rarely evaluated. The main reason potentially lies in the fact that training on resource-restricted IoT devices is still challenging, e.g., to implement and manage. Notwithstanding, it is imperative to investigate the \textit{training performance of distributed learning techniques} as, in most cases, the data yielded by IoT devices are sensitive such as in smart-home and smart-health applications, and should not leave the localized device.

\vspace{-0.3cm}
\subsection{Our Studies} 
\subsubsection{Learning Performance} This work is firstly to evaluate the training performance of distributed learning techniques under IoT settings with a focus on the popular Federated Learning (FL)~\cite{fedlearningMcMahan17,fed2,wu2020safa} and the recent Split Learning (SL)~\cite{gupta2018distributed,nopeeksplitNN}. 
In addition, we consider a new hybrid distributed learning framework, namely splitfed learning (SFL)~\cite{splitfed}, that explores unique benefits provided by each of the FL and SL. By recognizing that the current SFL designs~\cite{splitfed} are specific, we propose a generalized SFL (SFLG), which is more suitable and flexible for the IoT applications as it can still well fit when the IoT devices are large-scaled deployed. 

For each of these distributed learning techniques, we evaluate and compare their learning performance in an end-to-end manner. The learning performance is evaluated with various datasets and various settings, including (i) independent and identically distributed (IID) data, (ii) imbalanced data, and (iii) non-IID or skewed data resided locally to resemble the heterogeneous data distribution characteristics for IoT devices in real-world. This is the first work to comprehensively compare the learning performance of these promising distributed learning techniques with a concentration on the IoT setting. Considering the importance of communication efficiency in IoT settings, we have proposed techniques to reduce it in the SFLG.

\subsubsection{On-Device Overhead} 
Furthermore, there is no empirical study on the end-to-end evaluation of FL, SL, and the SFL on real-world IoT devices, e.g., Raspberry Pi, in terms of their implementation or execution overhead, such as communication cost, power consumption, and training time. Indeed, as highlighted in~\cite{chen2019deep}, there is a demand to understand the deep learning performances on resource-constrained IoT/edge device hardware like  Raspberry Pi~\cite{raspberryPi}. Experimental results with real-world IoT devices would be useful for practitioners when choosing suitable distributed learning techniques for deployment. Thereby, this work aims to take the first step of empirically and systematically evaluate the training of FL, SL as well as SFL in real-world IoT devices---there exist Raspberry Pi implementations on FL~\cite{openmind,zhang2020efficient} but not on other distributed learning techniques. 
% as they do are not to systematically evaluate and compare them.

\vspace{-0.3cm}
\subsection{Contributions}  Overall, this work is to make distributed learning being more suitable for resource-restrict (e.g., computation and communication restricted) IoT applications. The main contributions/results (\textit{last three} are new or have been renewed in comparison with our conference work~\cite{gao2020end}) of this work are summarized as follows:

\begin{enumerate}
    \item We are the first~\cite{gao2020end} to evaluate SL learning performance in terms of model accuracy and convergence under non-IID and imbalanced data distributions, and then compare it with a popular counterpart FL under the same settings. Our \textit{empirical} results---up to simulated 100 clients---demonstrate that SL exhibits better learning performance than FL under imbalanced data but worse than FL under (extreme) non-IID data, indicating that SL accuracy is also sensitive to the heterogeneous characteristics of the distributed data. (\autoref{sec:learningEvaluation})

    \item We then, based on two specific SFL variants, propose a generalized SFL (namely SFLG) for exploiting the advantages of each of SL and FL to complement their individual shortcomings and then evaluate it with the same settings of FL and SL for a fair comparison. The SFLG obviates the cumbersome sequential training process among IoT devices and utilizes the rich-resource in the server-side to expedite the training while still retaining its low computational overhead advantage. (\autoref{sec:learningEvaluation}) 
    % --- your point 3 is correct 

    \item We take the first step toward fair comparisons of {\it training} performance between FL and SL by mounting both on Raspberry Pi. We provide detailed performance overhead evaluations of training time, amount of memory used, amount of power consumed, communication overhead, peak power, and temperature to serve as a reference for practitioners. In addition, (i) effects of the number of split layers in the SL, (ii) effects of models with different complexities for both SL and FL are quantified and compared. The further IoT implementation of SFL corroborates its substantial training time reduction while still maintaining the low computation overhead inherited from the SL. (\autoref{sec:implementationEvaluation})
    
    \item We validate that SFLG achieves good flexibility to trade-off learning performance, scalability, and training time expedition when the IoT devices are large-scaled. In addition, techniques of reducing SL and SFLG communication overhead are proposed and experimentally validated to better fit the scenario given the IoT applications could be bottlenecked with communication. Source codes of this work are released for facilitating future research and deployment~\url{https://github.com/garrisongys/SplitFed}\footnote{Artifacts of our previous study~\cite{gao2020end} is at~\url{https://github.com/Minki-Kim95/Federated-Learning-and-Split-Learning-with-raspberry-pi}}. (\autoref{sec:optimization})

\end{enumerate}

The remainder of this paper is organized as follows: Section~\ref{sec:background} details background on distributed learning techniques. Section~\ref{sec:sflg} presents our generalized SFL (namely SFLG). In Section~\ref{sec:learningEvaluation}, we comprehensively evaluate SL, FL, and SFL under various heterogeneous data distributions for end-to-end comparisons. Section~\ref{sec:implementationEvaluation} mounts these three distributed learning techniques on Raspberry Pi to empirically evaluate and compare their implementation overhead. Section~\ref{sec:optimization} validates the advantages of SFLG in large-scaled IoT devices and further investigates the pragmatic techniques to reduce the communication overhead of SL and SFLG. Section~\ref{sec:Conclusion} concludes this work.

%====================================================================
%                   BACKGROUND
%======================================================
\section{Background}
\label{sec:background}

In this section, we provide a brief background on federated learning, split learning, and splitfed learning. 
\subsection{Federated Learning}
\begin{figure}[!ht]
\begin{subfigure}{.5\textwidth}
  \centering
   \includegraphics[trim={2cm 2cm 2cm 2cm},clip,width=0.9\linewidth]{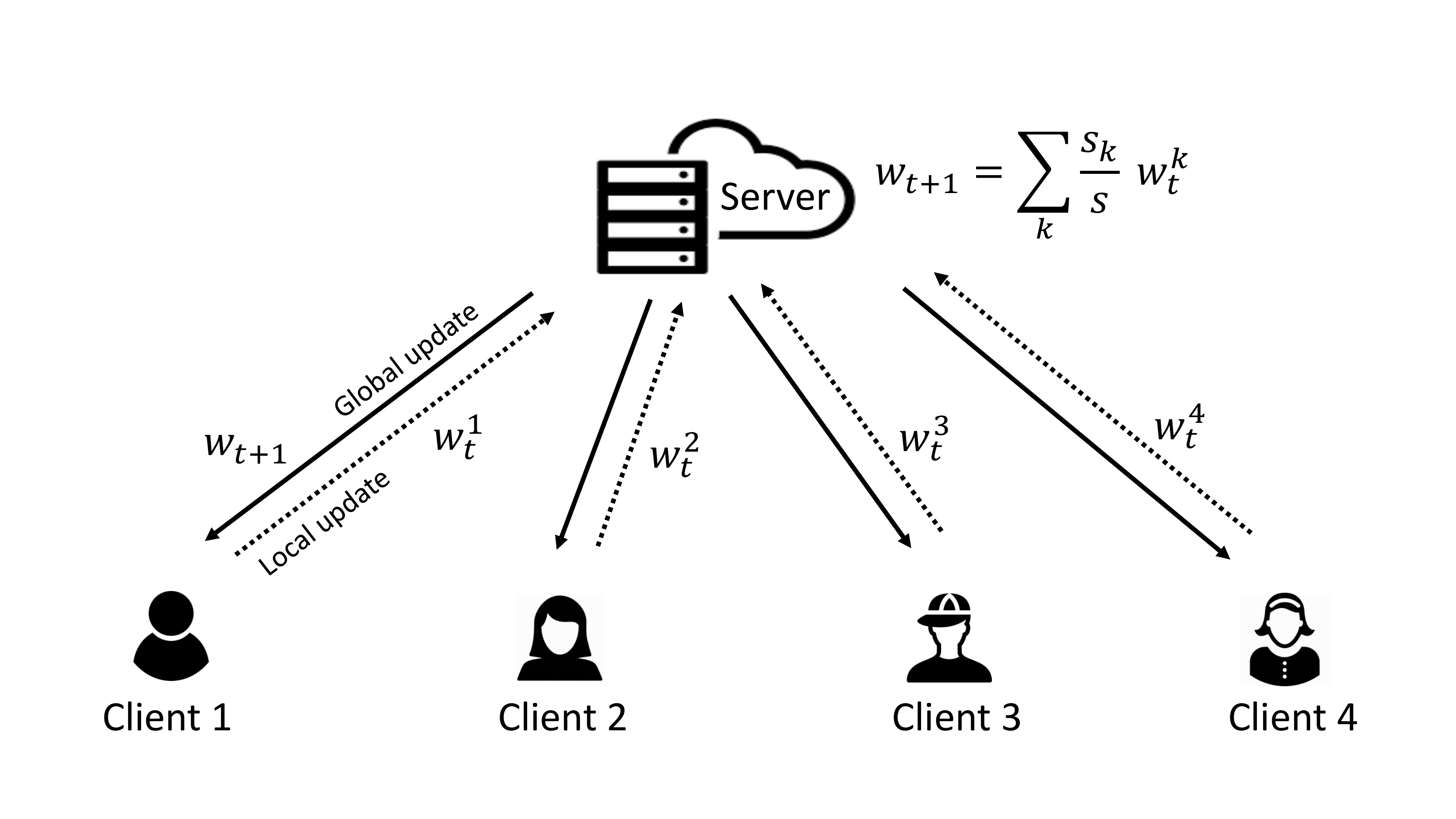}  
  \caption{Federated learning with four clients.}
  \label{fig:1}
\end{subfigure}
\begin{subfigure}{.5\textwidth}
  \centering
  \includegraphics[trim={0 2cm 0 0},clip=true,width=1\linewidth]{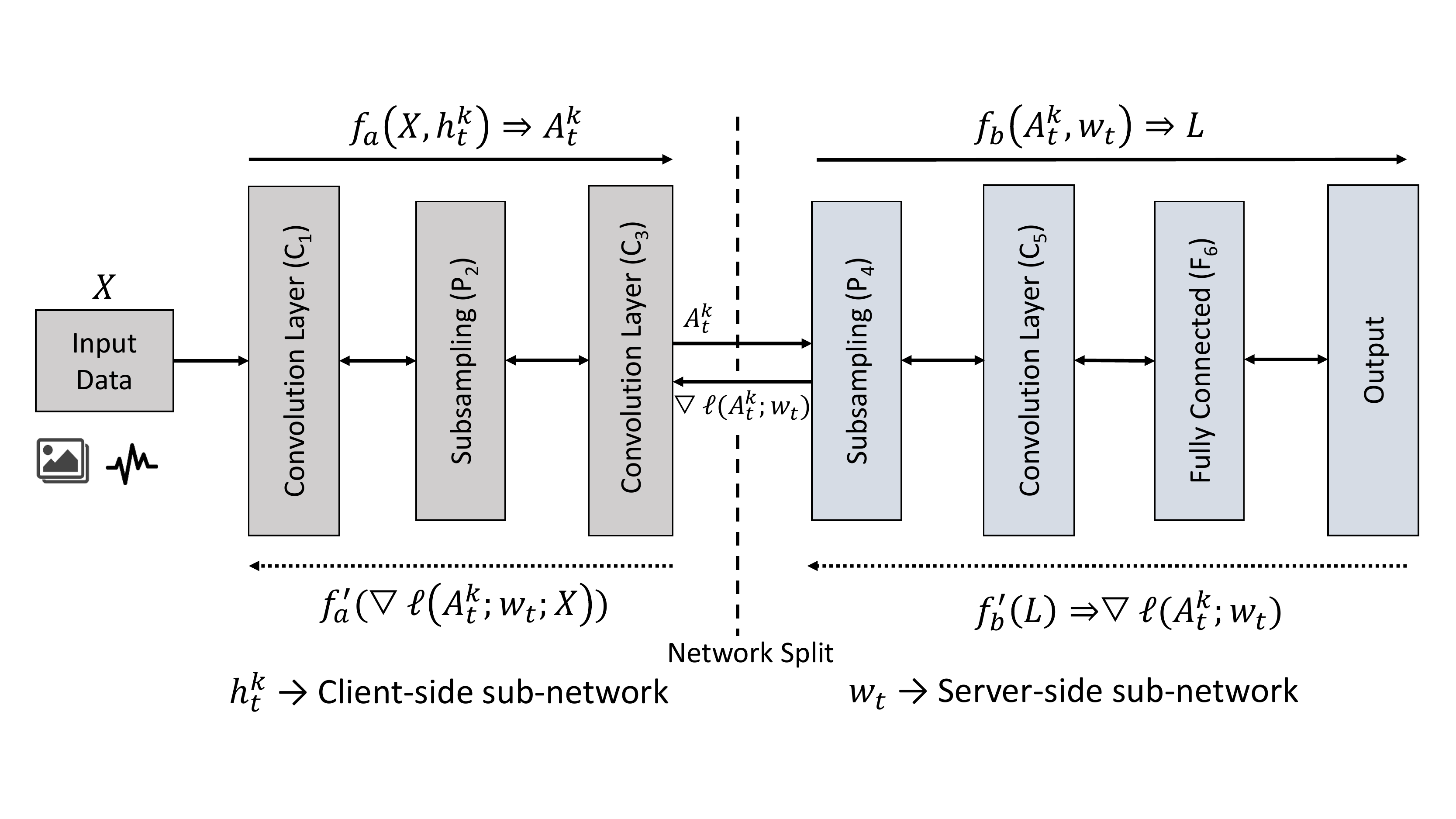}  
  \caption{Split learning with one client and server, where the global model has seven layers, and the model is split into two sub-networks, namely client-side sub-network (first three layers) and server-side sub-network (last four layers).}
  	\label{fig:2}
\end{subfigure}
\caption{Illustration of federated learning and split learning.}
\label{fig:fig}
\end{figure}

The federated learning (FL) is illustrated in Fig.~\ref{fig:1} where we focus on the representative FL learning with the average algorithm \texttt{FedAvg} for local model aggregation~\cite{mcmahan2016communication}. During the training process, the server first initializes the global model $w_t$ and sends it to all participating clients. After receiving the model $w_t$, each client $k$ trains the global model on its local data; $s_k$ is the number of training samples held by client $k$ while $s$ is the total number of training samples across all clients. Afterward, each client returns the locally updated model $w_t^k$ to the server. The server then aggregates all those models to update the global model $w_{t + 1}$. The above process (often called \emph{round}) repeatedly continues until the model converges. 

According to~\cite{mcmahan2016communication}, instead of training the model with local data only one epoch, each client trains the local model for several epochs before sending it to the server in one communication round, which is commonly used for FL optimization. Although \texttt{FedAvg} usually works well, specifically for non-convex problems, there are no convergence guarantees. FL may diverge in practical settings, especially if data are non-IID and imbalanced distributed across clients~\cite{li2019federated}.

\subsection{Split Learning}
% Split neural network (SL) or split learning is a distributed machine learning technique introduced recently to learn a joint model by keeping data decentralized. 
Unlike FL, in which each client trains the whole neural network, split learning (SL)~\cite{gupta2018distributed,nopeeksplitNN} divides a neural network-based model into at least two sub-networks and then trains the sub-networks, separately, on distributed parties (e.g., client and server). A simple example of SL is illustrated in Figure~\ref{fig:2}, where $C_3$ is the cut layer that divides the whole network into two sub-networks. The first sub-network $h_t$ is trained and accessed by the client; the second sub-network $w_t$ is trained and accessed by the server. 
While training, we have forward and backpropagation in the network. In Figure~\ref{fig:2}, the training occurs as follows: The client carries forward propagation over the input data, then sends the activations of the cut layer, called smashed data ($A_t^k$), to the server. The server carries forward propagation over the smashed data and calculates loss. Then, backpropagation starts over the loss, and yields the gradients of the smashed data ($\triangledown \ell(A_t^k;w_t)$) during the process. Afterward, the gradient is sent to the client, which then carries its backpropagation. 
In SL training/testing, the server has no access to clients' sub-networks and data, which provides privacy. Besides the privacy benefit, each client only needs to train a sub-network usually consisting of a few layers while most layers reside in the server. Therefore, as the other benefit, the client's computation load is reduced.

The learning performance (e.g., model accuracy and convergence) of SL has not been investigated yet when the data is non-IID or distributed in an imbalanced manner, which has been considered. 

\subsection{Splitfed learning}\label{sec:splitfed}
SL greatly reduces the computation requirement on the client-side as it only computes on a (small) sub-network. However, it needs to sequentially iterate over each client, which results in prolonged training time if multiple clients are present. 
In FL, each client usually interacts with the server in parallel; thus, training can be completed faster than SL. However, each client has to train the entire model that renders high computational overhead. 

Recently, SL and FL are blended to take advantage of both. It is called splitfed learning (SFL)~\cite{splitfed}. 
In SFL, all clients compute in parallel and independently. They send/receive their smashed data to/from the server in parallel. The client-side sub-network synchronization, i.e., forming the global client-side network, is done by aggregating (e.g., weighted averaging) all client-side local networks in a separate server, called fed server. Two ways of server-side sub-network synchronization are presented in~\cite{splitfed}, which correspondingly results in two specific variants of SLF. 

\vspace{2pt}\noindent$\bullet$ {\bf SFLV1:} Firstly, by performing parallel and independent training over the smashed data of each client, which \textit{results in the number of server-side sub-networks equals the number of clients.} Later, all the sub-networks are aggregated (e.g., weighted averaged) to form the global server-side network. This is referred to as SFLV1. 

\vspace{2pt}\noindent$\bullet$ {\bf SFLV2:} Secondly, by performing sequential server-side sub-network training over the smashed data of each client---note the client can still send their smashed data concurrently. \textit{This keeps only one copy of the sub-network on the server-side}, and it is the global server-side network once the main server processed over all smashed data. This is referred to as SFLV2.   

By observing the possibility of merging these two specific SFLV1 and SFLV2 algorithms, this work generalizes them. The generalized algorithm enables a varying number of server-side sub-networks other than that of SFLV1 and SFLV2 in the main server-side while training. Thus one can flexibly choose the number based on the available server-side computational capacity, which is suitable for large-scale IoT devices focused by our study. 
% We consider the generalized SFL in our analysis in this paper.

\section{Generalizing Splitfed Learning}\label{sec:sflg}

\begin{figure}[h] 
	\centering
	\includegraphics[trim=6cm 2.1cm 4.3cm 3.2cm,clip, width=0.5\textwidth]{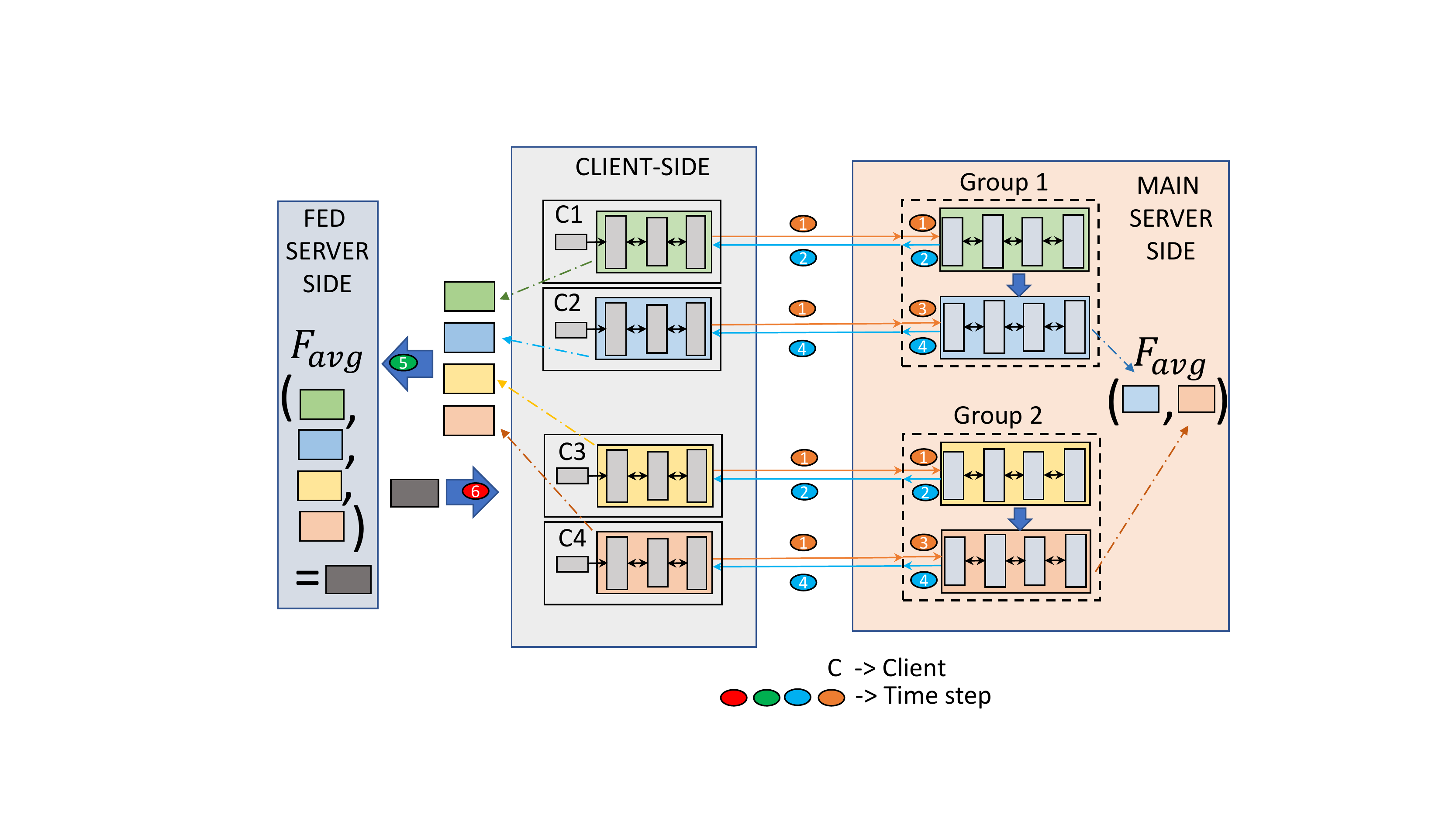}
	\caption{Illustration of the generalized splitfed learning (\mbox{SFLG}) by an example with four clients (C1, C2, C3, and C4), and \textit{two groups at the main server-side, where each group is handling the operations with two clients}. The server-side model training inside each group is sequential, and the inter-group is in parallel. The server-side global model and the client-side global model are formed by the weighted averaging ($F_{avg}(.)$).}
	\label{fig:splitfed_general}
\end{figure}

In the generalization of SFL, the client-side operations, including client-side model synchronization, are kept the same. \textit{The change is only in the server-side training operations}. In the generalised SFL, we call \emph{SFLG}, firstly, the server considers several $G$ groups, where each group has several clients (here client means clients' identifications). In a simple setup, the clients in a group are unique, and no two groups share the same client. Then, the server-side model is trained and updated sequentially within the group (only one copy of the server-side model is kept), whereas the group operates independently and in parallel. Thus, at the end of the executions of the server-side operations, we get the $|G|$ number of trained server-side models. Finally, these models are aggregated (e.g., weighted averaging) to make the server-side global model. This server-side model training process is depicted in Figure.~\ref{fig:splitfed_general}, while details are referred to Algorithm~\ref{algo:SFLG}. Though this algorithm is presented with the label sharing case, it is possible for various other configurations, such as without label sharing, extended split, and vertically partitioned data, and it is performed similarly to the description provided in split learning~\cite{split_differentconfiguration,splitbook}.  

\begin{algorithm}  
	\scriptsize
	\SetNoFillComment
	\caption{Generalized splitfed learning (SFLG) with label sharing.}
	%------------------------------------------------------------------------------------------------------
	\SetAlgoNoLine	
	\textbf{Notations:}\\
	$S_t$: A set of participating clients at time $t$.\\  
	$G$: A set of sets of clients (from $S_t$) representing groups at the server-side. \\
	$n_{g}$: The sum of the number of samples of the clients present in a group $g \in G$.\\ 
	$n_{k}$: The number of samples at client $k\in \{1,2,\dotsc,K\}$.\\ 
	$n$: Total number of samples considering all clients.\\
	$X^k$: Data in client $k$.\\
	$\eta$: Learning rate fixed both at the client and server-side.\\
	$t$: Time instance.\\
	
	\vskip8pt
	\SetKwProg{Fn}{MainServer executes:}{}{} \tcc{	\scriptsize Runs on Main Server}
	\Fn{} {
	    	\If {$t$=0}{
	    Initialize $w_{t}$ (global server-side model)\\
	    }
	    \For{\textup{each group $ g\in G $ in parallel}} {
	        $w_t^{g} = w_t$\\
	        \For{\textup{each client $ k\in g $ in serial}} {
	            $w_t^k \leftarrow w_t^{g}$\\
	            \While{local epoch $i \neq E$}{
	              	$ (A_{t,i}^k, Y^k_{t,i})  \leftarrow$ ClientForwardPropagation$(h_{t,i}^k)$ \\
    			    Forward propagation: Compute $f_b(A_{t,i}^k, w_{t}^k)$ to calculate $\hat{Y}^k_{t,i}$, then calculate loss $L$ from $Y^k_{t,i}$ and $\hat{Y}^k_{t,i}$\\
    	            Back-propagation: Compute $f'_b(L)$ yielding gradients $\triangledown \ell (w_{t,i}^k)$ and $\triangledown \ell (A_{t,i}^k;w_t^k)$\\
    		        
    		        Update: $w_{t}^k\leftarrow w_{t}^k -\eta \triangledown \ell (w_{t,i}^k)$\\
    		    	 
    		    	 Update: $w_t^{g} \leftarrow w_t^k$\\
    		    	
    		    	Send the gradient $\triangledown \ell (A_{t,i}^k; w_{t}^k)$ of the cut layer to client $k$ for ClientBackPropagation$(\triangledown \ell (A_{t,i}^k; w_{t}^k))$ 
    		    	}
    	 }
    	 }
			$ w_{t+1} \leftarrow \sum_{g} \frac{n_g}{n} w_{t}^g$
	}
	%-----------------------------------------------------
	\vskip8pt
	\SetKwProg{Fn}{ClientForwardPropagation  ($h_{t,i}^k$) :}{}{} \tcc{	\scriptsize Runs on Client $k$}
	\Fn{} {
	    Update $h_{t}^k  \leftarrow$ FedServer()\\
	    Set $A_t^k =\phi$\\
	    $h_{t,i-1}^k \leftarrow h_{t}^k$\\
		    \For{\textup{each local epoch $i$ from $1$ to $E$}}{
		            $ h_{t,i}^k \leftarrow h_{t,i-1}^k$\\
    				Forward propagation: calculate $f_a (X^k, h_{t,i}^k)$ yielding
    				the activations of the final layer $A_{t,i}^k$\\
    				The true labels of $X^k$ is $Y^k_{t,i}$
    			
    			Send $A_{t,i}^k$ and $Y^k_{t,i}$ to the main server\\ 
    			Wait for the completion of ClientBackPropagation($\triangledown \ell (A_{t,i}^k, w_{t,i}^k)$) where $h_{t,i}^k$ is updated
    		}

	}
	%------------------------------------------------
    \vskip8pt
	\SetKwProg{Fn}{ClientBackPropagation($\triangledown \ell (A_{t,i}^k, w_{t,i}^k)$):}{}{} \tcc{\scriptsize Runs on Client $k$}
	\Fn{} {
	    
	    \For{\textup{each local epoch $i$ from $1$ to $E$}} {
	    Receive gradient of the cut layer: $\triangledown \ell (A_{t,i}^k, w_{t,i}^k) \leftarrow$~MainServer()\\
	       
		        Back-propagation: compute $f'_a (\triangledown \ell (A_{t,i}^k; w_{t,i}^k; X^k))$ yielding $\triangledown \ell (h_{t,i}^k; X^k))$\\

		    Update $h_{t,i}^k \leftarrow h_{t,i}^k -\eta \triangledown \ell (h_{t,i}^k; X^k))$
		    
	    }
	    Send $h_{t,E}^k$ to Fedserver() 
	}
	%------------------------------------------------
	\vskip8pt
	\SetKwProg{Fn}{FedServer executes:}{}{}	 \tcc{	\scriptsize Runs on Fed Server}	
		\Fn{}{
		\eIf {t=0}{
		Initialize $h_{t}$ (global client weights)\\
		Send $h_t $  to all $K $ clients for ClientForwardPropagation$(h^{k}_{t,i})$
		}{
		    \For{\textup{each client $ k \in  S_t$ in parallel}}{
				$ \mathbf{h}^{k}_{t} \leftarrow$ ClientBackprop($\triangledown \ell (A_{t,i}^k, w_{t,i}^k)$) 
			}
			Client-side global model updates: $ h_t  \leftarrow \sum_{k=1}^{K} \frac{n_k}{n}  h^{k}_t$
			
		    Send $h_t $  to all $K $ clients for ClientForwardPropagation$(h^{k}_{t,i})$	
			
		}
	}
	\label{algo:SFLG}
\end{algorithm}

By observing Algorithm~\ref{algo:SFLG}, it is not difficult to see the following:
\begin{preposition}
(i) If group $|G|=n$ at the server-side and each group includes a unique client, and no two groups include the same client, then SFLG is SFLV1. (ii) If group $|G|=1$ at the server-side and that group includes all participating clients, then SFLG is SFLV2. 
\end{preposition}

This preposition will be validated in Section~\ref{sec:optimization}. The straightforward benefit of SFLG, including SFLV1 and SFLV2 variants, is that the client-side training can be performed in parallel for the expedition, consequentially saves power---waiting is no longer required by setting the IoT devices into idle status. Moreover, one research challenge---the worsen accuracy performance under non-IID that is recognized in our previous study~\cite{gao2020end}---faced by the SL can be mitigated.

%====================================================================
%                   Experimental setup
%====================================================================
% \section{Experimental setup}

%==============================================================
\section{Learning Performance Evaluation and Comparison}\label{sec:learningEvaluation}

\subsection{Datasets and Models}
% \sharif{We can follow figure \ref{fig:evaluationOverview} here by saying we experiment with two types of datasets: image e.g. CIFAR10 to be fed into 2D CNN and time-series sequential data e.g. Speech commands and ECG to be fed into 1D CNN}

Sequential data or time-series data is pervasively collected and processed by IoT devices. For example, people can order and purchase goods using speech commands at their voice assistant. Wearable medical sensors are used to monitor users' health status in real-time. Consequentially, we choose two such popular datasets: speech command (SC) and ECG for experimental evaluations, as summarized in Table~\ref{tab:Setup}. The SC is a personalized dataset, and the ECG is a medical dataset. Both datasets would be privacy-sensitive, where users are unwilling to share. Regarding model architecture, the 1D CNN has been recently shown to be efficient for dealing with sequential data~\cite{kiranyaz20191d}, which we have adopted. As a matter of fact, the major consideration of 1D CNN usage instead of sequential machine learning models such as LTSM and RNN for sequential data is because there is no effective solution for splitting the sequential model in the SL setting. The SL is currently only applicable to vertical model architectures, in particular, the CNN.

Notably, we have indeed considered image classification tasks and 2D CNN models. Although Raspberry Pi is regarded as a high-end IoT device~\cite{musaddiq2018survey}, its computational resources are still quite limited compared with traditional computing devices such as servers and even PCs. We first tried to run MobileNet~\cite{howard2017mobilenets}. When we trained the CIFAR10 dataset~\cite{krizhevsky2009learning} with MobileNetv1\footnote{Source code is adopted from \url{https://github.com/Tshzzz/cifar10.classifer/blob/master/models/mobilenet.py}.} (20 conv2D layers with 3,228,170 model parameters in total), it took 8 hours 41 minutes for FL per round with 1 local epoch. For SL, it took about 2.5 hours for one epoch across five Raspberry Pi devices\footnote{Since the training sample for CIFAR10 is 50,000, we use 5 clients. Therefore, each client holds 10,000 images for both FL and SL. We run it in default without delicate optimization.}, when only the first two layers are running on the Raspberry Pi device. In addition, we tried to run ResNet20\footnote{Source code is adopted from \url{https://github.com/akamaster/pytorch_resnet_cifar10/blob/master/resnet.py}.} (with 20 conv2D layers and 269,722 parameters). When we ran SL across 5 Raspberry Pi devices, it took about 1 hour to finish one round when the first two layers are only running on the Pi devices. When we also ran FL across 5 Raspberry Pi devices, it took 37 minutes for one round with one local epoch. 

Therefore, this work mainly uses two sequential datasets and 1D CNN models, summarized in Table~\ref{tab:Setup}, for evaluation and comparison.

\subsubsection{Speech Commands (SC)} This task is for speech command recognition. The SC contains many one-second '.wav' audio files: each sample has a single spoken English word~\cite{SClink}. These words are from a small set of commands and are spoken by a variety of different speakers. In our experiments, we use 10 classes: `zero', `one', `two', `three', `four', `five', `six', `seven', `eight', and `nine.' There are 20,827 samples where 11,360 samples are used for training, and the remaining samples are used for testing.

% We use the 1D CNN model with 5 CNN layers and 3 dense layers. 

\subsubsection{Electrocardiogram (ECG)} 
MIT-BIH arrhythmia \cite{moody2001impact} is a popular dataset for ECG signal classification or arrhythmia diagnosis detection models. Following~\cite{kiranyaz2015real, li2017classification}, we collect 26,490 samples in total which represent 5 heartbeat types as classification targets: $N$ (normal beat), $L$ (left bundle branch block), $R$ (right bundle branch block), $A$ (atrial premature contraction), and $V$ (ventricular premature contraction). Half of them are randomly chosen for training, while the rest samples are for testing.

\begin{table}
	\centering 
	\caption{Datasets and Models.}
			\resizebox{0.5\textwidth}{!}{
	\begin{tabular}{c| c | c | c | c | c | c} % 
		\toprule % Top horizontal line
		\toprule % Top horizontal line
				
		Dataset &  \begin{tabular}{@{}c@{}} $\#$ of  \\ labels \end{tabular}  & \begin{tabular}{@{}c@{}} Input  \\ size\end{tabular} & \begin{tabular}{@{}c@{}} $\#$ of  \\ samples \end{tabular} & \begin{tabular}{@{}c@{}} Model  \\ Architecture \end{tabular} & \begin{tabular}{@{}c@{}} Total  \\ Parameters \end{tabular} & \begin{tabular}{@{}c@{}} Total Model Accuracy \\ (Centralized data)  \end{tabular} \\ % Column names row
		\midrule
		ECG &  5 & 124 &  26,490 & \begin{tabular}{@{}c@{}} 4conv + 2dense \\ 1D CNN \end{tabular} & \begin{tabular}{@{}c@{}} 68,901 \end{tabular} & 97.78\%\\ 
		\hline
		\begin{tabular}{@{}c@{}} Speech Command \\ (SC) \end{tabular}  &  10 & 8,000 &  32,187 & \begin{tabular}{@{}c@{}} 4conv + 2dense \\ 1D CNN \end{tabular} & \begin{tabular}{@{}c@{}} 522,586 \end{tabular} & 85.29\% \\ 
		\bottomrule
	\end{tabular}
			}
	\label{tab:Setup} % A label for referencing this table elsewhere, references are used in text as \ref{label}
\end{table}

\subsection{Data Distribution Considerations}
In practice, data is often distributed among clients in an imbalanced manner, e.g., some sensors are more active than others---with more data, and non-IID distributed, e.g., a single person's data can only be collected~\cite{li2019federated}. Going forward, we evaluate the learning performance of FL, SL, and SFL under the same data distribution setting, including IID, imbalanced, and non-IID for quantitative comparisons. In all cases, the data distributed to each client are non-overlapped. 

As for the SFL, herein we focus on the SFLV1 (that is equal to set $|G|=n$ of the SFLG) as it gains the maximum parallelization capability, which ultimately outsources and reaps the rich computation in the server. This is always preferable, in particular, when the number of participants is not that large. We defer comprehensive learning comparisons among our generalized SFLG with specific SFLV1 ($|G|=n$ of the SFLG) and SFLV2 ($|G|=1$ of the SFLG)~\cite{splitfed} to demonstrate the SFLG flexibility in large-scaled IoT devices in~\autoref{sec:SFvariant}.

\subsection{IID and Balanced Dataset}\label{sec:idealDist}
Starting with ideal IID and balanced data distribution, we evaluate FL, SL, and SFLV1 through both SC and ECG datasets\footnote{For all tests in this section if there is no explicit statement, the 4conv+2dense 1D CNN model architecture is used. The learning rate is set to be 0.001. The batch size = 32.}

\begin{figure}[t]
	\centering
	\includegraphics[trim=0 0 0 0,clip,width=0.5\textwidth]{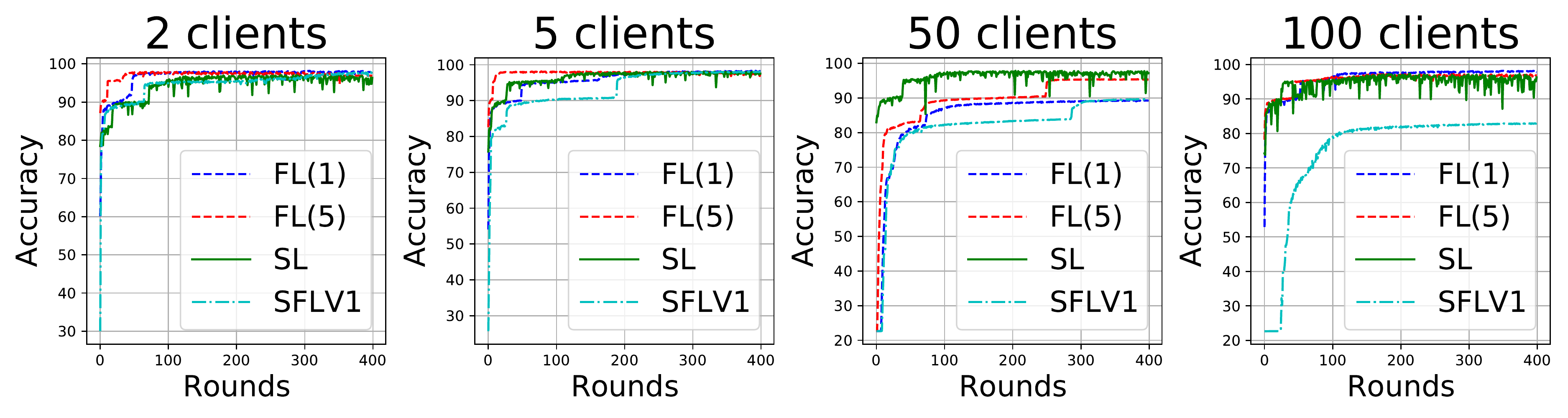}
	\caption{Testing accuracy of FL (1 and 5 local epochs for 1 round), SL, and SFLV1 over rounds for the ECG data, which is IID and distributed in a balanced manner.}
	\label{fig:IID_multiple_clients}
\end{figure}

\begin{figure}[t]
	\centering
	\includegraphics[trim=0 0 0 0,clip,width=0.5\textwidth]{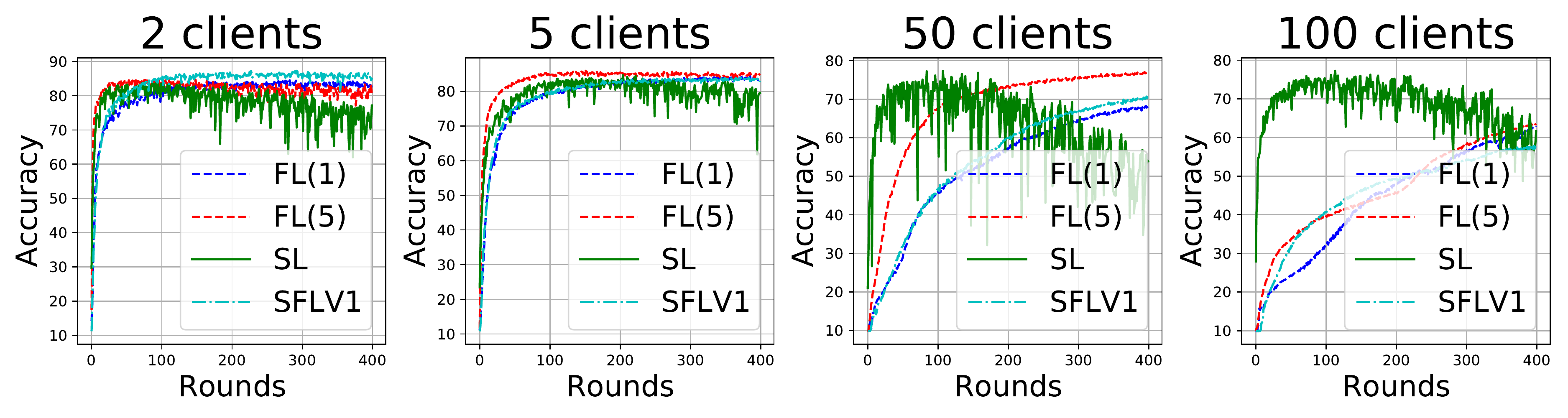}
	\caption{Testing accuracy of FL (1 and 5 local epochs for 1 round), SL, and SFLV1 over rounds for the SC data, which is IID and distributed in a balanced manner.}
	\label{fig:IID_multiple_clients_sc}
\end{figure}

Fig.~\ref{fig:IID_multiple_clients} and~\ref{fig:IID_multiple_clients_sc} detail the testing accuracy as a function of the number of rounds when FL, SL, and SFLV1 are trained by a different number of clients---2, 5, 50, and 100 clients. We can see that SL can always converge relatively faster than FL with one local epoch---notably for SL and SFLV1, the local epoch can only be 1. FL struggles with the convergence, especially when the number of clients becomes large. 


The testing accuracy exhibits a drop after reaching an optimum point (especially the SC dataset in Fig.~\ref{fig:IID_multiple_clients_sc}). Thus, an increasing number of rounds appears not to help to improve accuracy. Stopping training at the optimal point saves training time. In addition, SL always exhibits an unstable learning curve with a high number of spikes. 

Furthermore, SL's model accuracy cannot reach the baseline accuracy of the centralized model---85.29\% for the SC and 97.78\% for the ECG, as detailed in Table~\ref{tab:Setup}. This limitation is clearly shown in Fig.~\ref{fig:IID_multiple_clients_sc}, when the number of clients is 50 or 100.

These results indicate that the SL model accuracy and convergence performance are not always the same as that of training a model through centralized data. Our findings are consistent with the previous conclusion in~\cite{gupta2018distributed}. However, we note that our findings are more generalized because we do not assume that the order of the data that arrived at multiple entities should be preserved, and the same initialization is used for assigning weights.

As for the SFLV1, under expectation, its performance approaches FL. This is because, from the model aggregation perspective, the SFLV1 is similar to FL. The only difference is that now the local model is not fully trained on the client but split by the IoT device and the server to reduce the computational overhead on-device.

\begin{mdframed}[backgroundcolor=black!10,rightline=false,leftline=false,topline=false,bottomline=false,roundcorner=2mm]
	\textbf{Notes:} (i) The SL learning performance is affected by the number of clients that is consistent with the previous conclusion in~\cite{gupta2018distributed}. (ii) The SL always outperforms the FL in terms of convergence speed in our experiments. Its training process exhibits unstable spikes, and the testing accuracy turns down once it reaches the optimal. (iii) The SFLV1 performance is close to the FL when the number of local epochs is set to be 1.
\end{mdframed}

\subsection{Imbalanced Data Distribution}\label{sec:imbaDist}
\begin{figure}[t]
	\centering
	\includegraphics[trim=0 0 0 0,clip,width=0.5\textwidth]{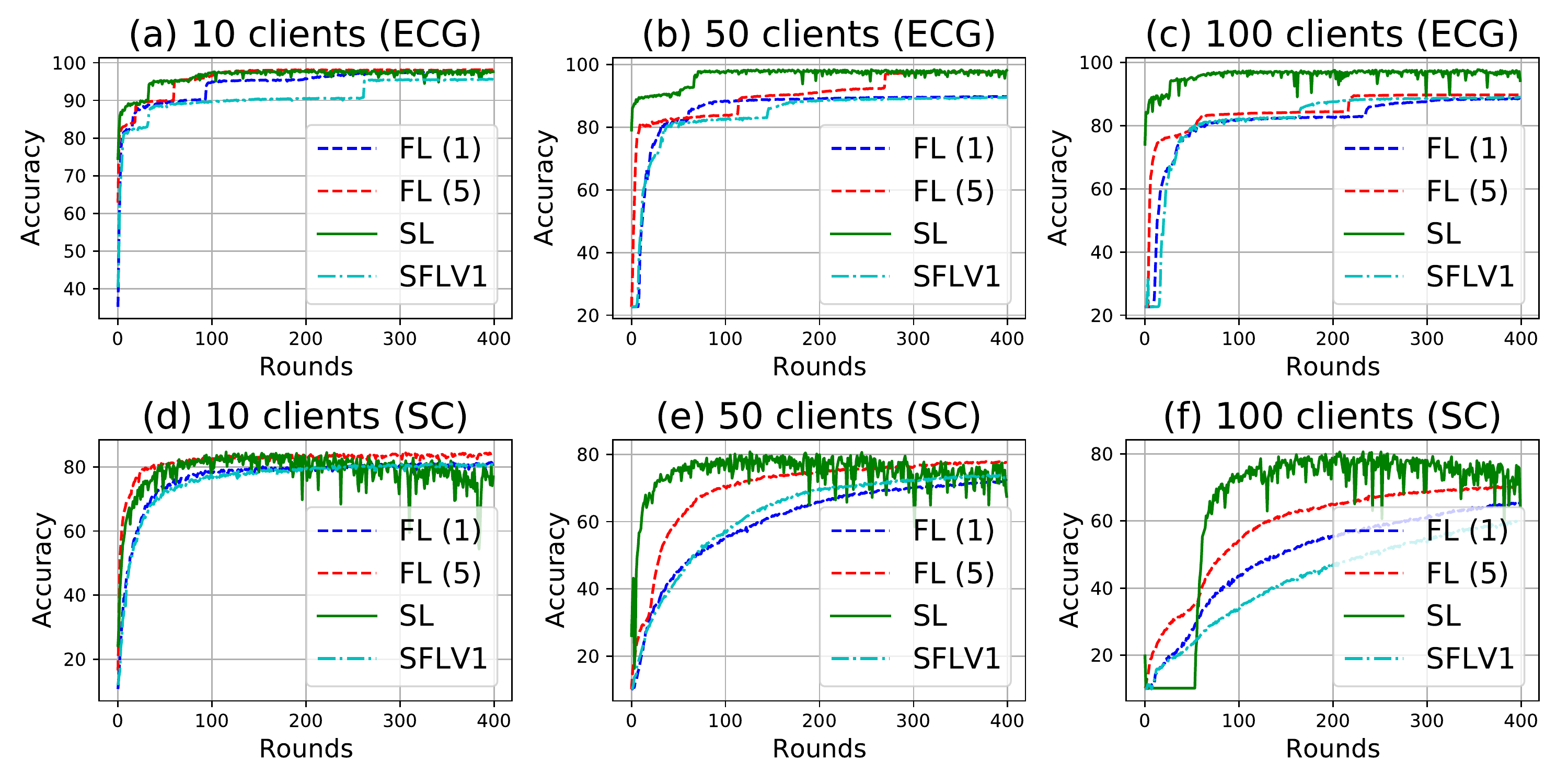}
	\caption{Testing accuracy of FL (1 and 5 local epochs for 1 round), SL, and SFLV1 over rounds with imbalanced data setting.}
% 	Top subfigures are for the ECG dataset while bottom subfigures are for the SC dataset.}
	\label{fig:Imbalance}
\end{figure}

We assume the data are distributed among clients following the {\it normal distribution} to simulate the realistic imbalanced data distribution. Larger the sigma/variance, the more imbalanced the data distributed. For example, when the number of clients is 10, and the total number of SC training dataset is 11360, the minimum number of training samples held by one client could be as few as 48 while the maximum number of training samples held by a client could be 3855---this is the setting for Fig.~\ref{fig:Imbalance} (d). We simulated clients up to 100. Given the same number of clients, {\it same data distribution} is applied to FL and SL and SFLV1.

According to Fig.~\ref{fig:Imbalance}, FL is hard to achieve the baseline accuracy of the centralized model, even when multiple local epochs per round is adopted for a large number of clients. For the SL, its model accuracy also deteriorates when the number of clients is large, e.g., Fig.~\ref{fig:Imbalance} (e) and (f). In addition, FL converges slower, especially when the number of clients goes up, e.g., 50 and 100 cases. Usage of more local epochs per round can expedite the convergence issue---but it cannot wholly prevent---{\it given the similar communication overhead}. However, more local epochs will proportionally prolong training time, consequentially more power consumption, on the IoT device, although it can reduce the communication overhead. SL is less sensitive to imbalanced data distribution since it always demonstrates a fast converge. In Fig.~\ref{fig:Imbalance} (f), we can see that the training of SL does not learn for the first 50 rounds/epochs. Once it starts learning, it indeed finds convergence quickly. As for the SFLV1, its learning performance closely approaches FL.

\begin{mdframed}[backgroundcolor=black!10,rightline=false,leftline=false,topline=false,bottomline=false,roundcorner=2mm]
	\textbf{Notes:} SL learning performance is affected by both the number of clients and imbalanced data distribution. In most cases, SL converges faster than FL in our experiments. The SFLV1 learning performance closely approaches FL. 
\end{mdframed}

\subsection{Non-IID Data Distribution}\label{sec:nonIIDDist}

\begin{figure*}
	\centering
	\includegraphics[trim=0 0 0 0,clip,width=0.85\textwidth]{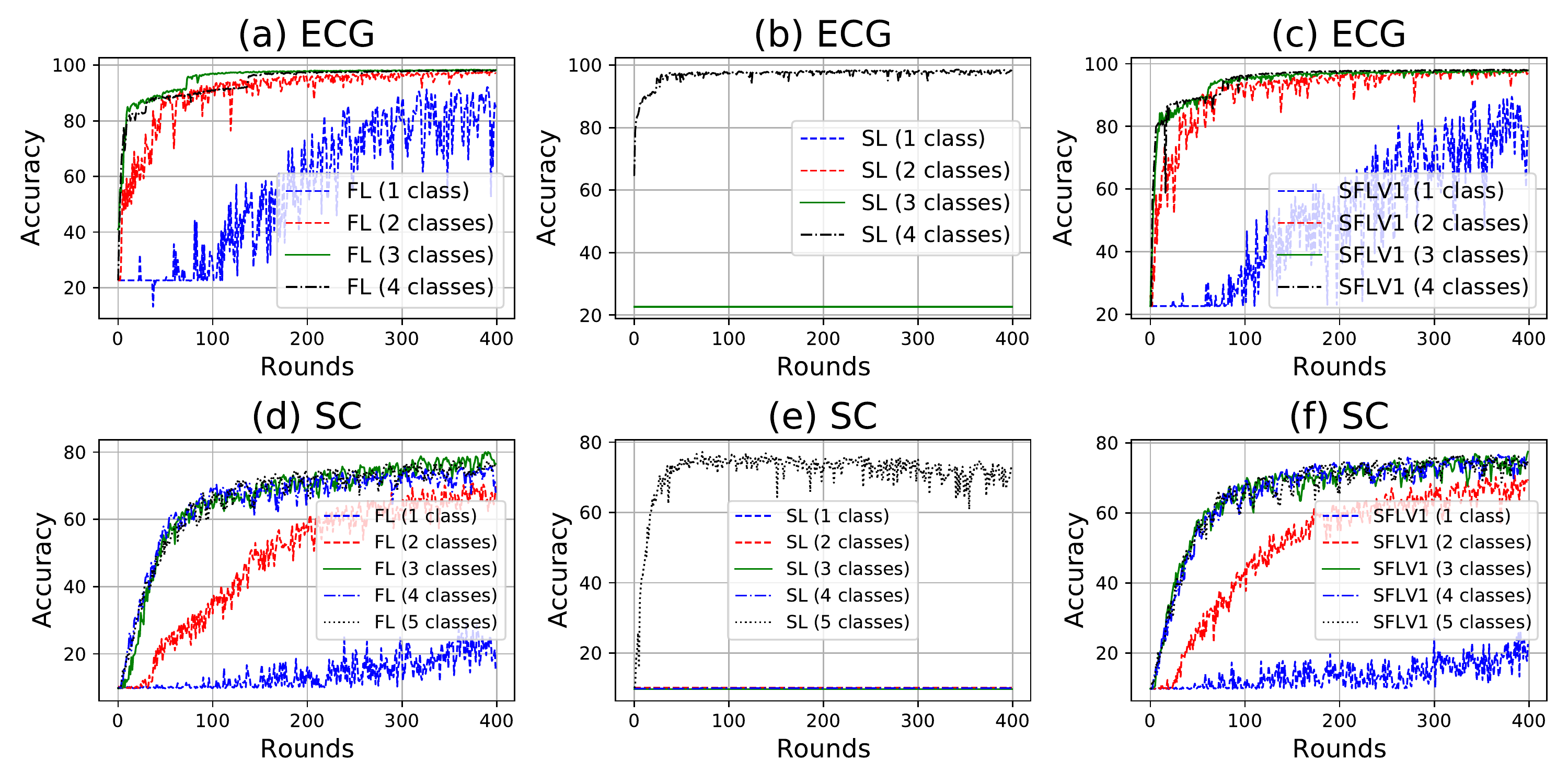}
	\caption{Testing accuracy of FL (1 and 5 local epochs for 1 round), SL, and SFLV1 over rounds with Non-IID dataset setting. (\# class(es)) refers that each client is with data from \# number of class(es).} 
	\label{fig:nonIID}
\end{figure*}

For the non-IID setting, the SC and ECG datasets are first sorted by class. Each client then receives data partition from only one single class, two classes, three classes, four classes, and five classes.

Results are detailed in Fig.~\ref{fig:nonIID}. Under extreme non-IID case (each client has only one class data), both SL and FL face convergence difficulties FL convergence significantly. Other than 1 client on a class case, the FL can always learn to converge better. For the SL, we can see that its learning almost die. In fact, we found that the predicted label all go to a single class.

\begin{mdframed}[backgroundcolor=black!10,rightline=false,leftline=false,topline=false,bottomline=false,roundcorner=2mm]
	\textbf{Notes:} SL is very sensitive to highly skewed data distribution. In fact, for both FL and SL, some extend of knowledge forgetting while learning is evident when trained in the non-IID settings. Moreover, in our experiments, FL outperforms SL under non-IID data settings, especially in extreme cases.  Again, the SFLV1 learning performance under non-IID is approaching FL.
\end{mdframed}
The possible reason for FL with better learning performance under extreme non-IID data lies in the approach of model training. FL aggregates (averages) the local models trained on the local data present at each client. The aggregated model usually has more knowledge of the data classes even though there is one class per client than the cases where the model is sequentially learned over the clients without aggregation, such as in SL.

\section{Implementation Overhead Evaluation on Raspberry Pi}\label{sec:implementationEvaluation}

\begin{figure}[h] % 
	\centering
	\includegraphics[trim=0 0 0 0,clip,width=0.48\textwidth]{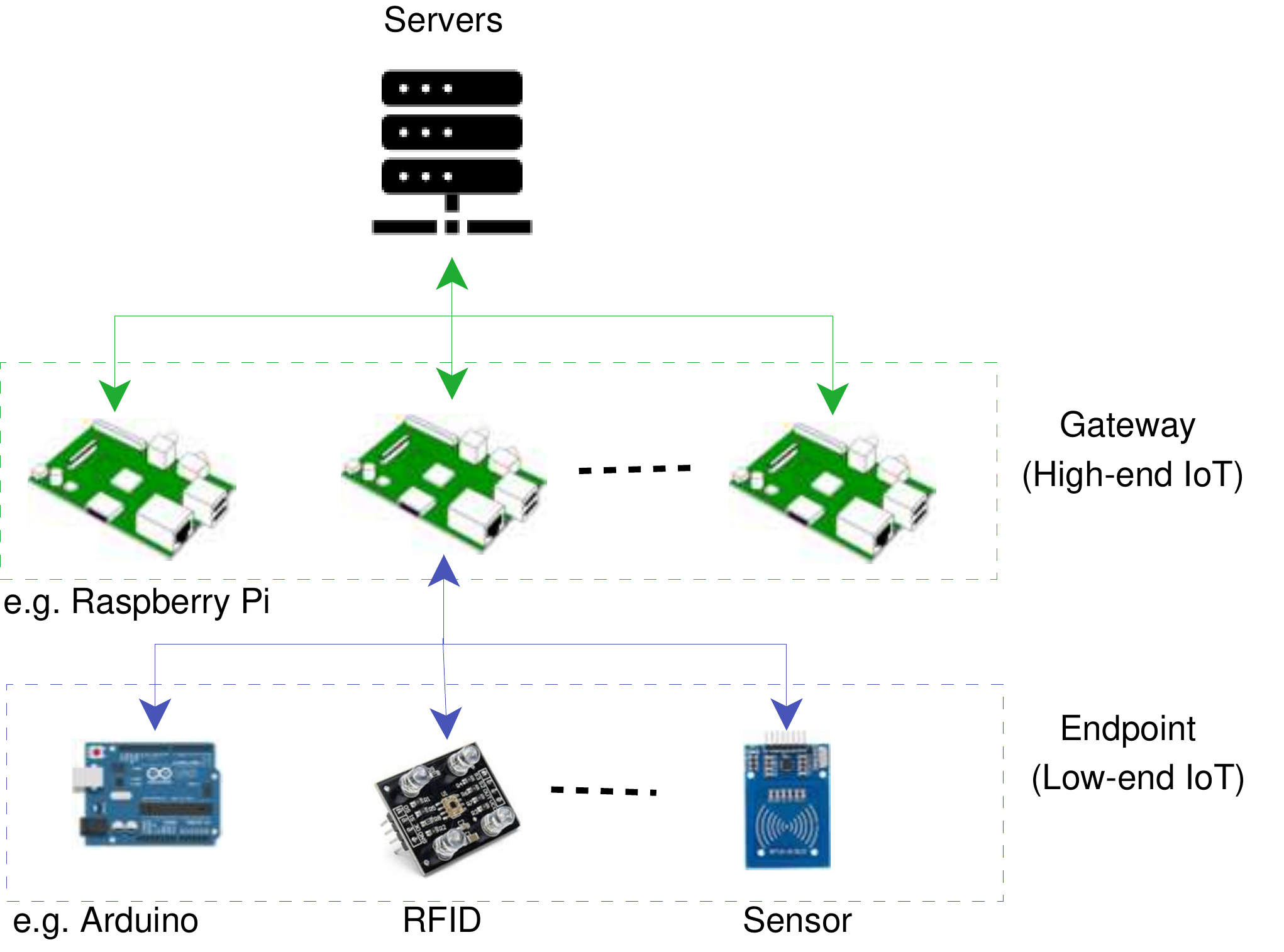}
	\caption{A typical IoT application setting. The IoT gateway (e.g., Raspberry Pi) aggregates data (e.g., from various IoT sensors) and interacts with the server to perform distributed learning.}
	\label{fig:IoTGateway}
\end{figure}

% {\bf Surya - I want this section to be organised in the same way like before. List the research questions and use each sub-section to address the research question. At the moment, I felt like you are generating the research questions after doing the experiments (that might be true in reality), but while presenting in the paper we should put RQs and do the experiments to answer them. }
Using the ECG dataset, we evaluate various overhead metrics such as time, power, communication, and memory when running FL, SL, and SFLV1 on Raspberry Pis that are representative IoT devices to provide a benchmark under real-world IoT settings. In particular, we simulate one typical IoT application scenario, as illustrated in Fig.~\ref{fig:IoTGateway}, similar to~\cite{nguyen2019diot}, which can be a smart home setting. According to~\cite{musaddiq2018survey}, the IoT device can be generally categorized to \emph{high-end} IoT device and \emph{low-end} IoT device. The low-end IoT devices are temperature, motion sensors, and RFID cards, which are usually strictly resource-constraint. They may not even support an OS such as Linux to run a machine learning algorithm. High-end IoT devices are simple devices like Raspberry Pi. Hence, in this simulated IoT application scenario, Pi serves as a gateway, which aggregates data from low-end IoT devices, e.g., sensors, and interacts with the server to perform distributed learning tasks. 

\subsection{Experimental Setup}
% \subsubsection{Scenario Stimulation}
% \subsection{}
% Each Raspberry PI acts as a client and the laptop acts as the server. 

\begin{figure}[!ht]
	\centering
	\includegraphics[trim=0 0 0 0,clip,width=0.5\textwidth]{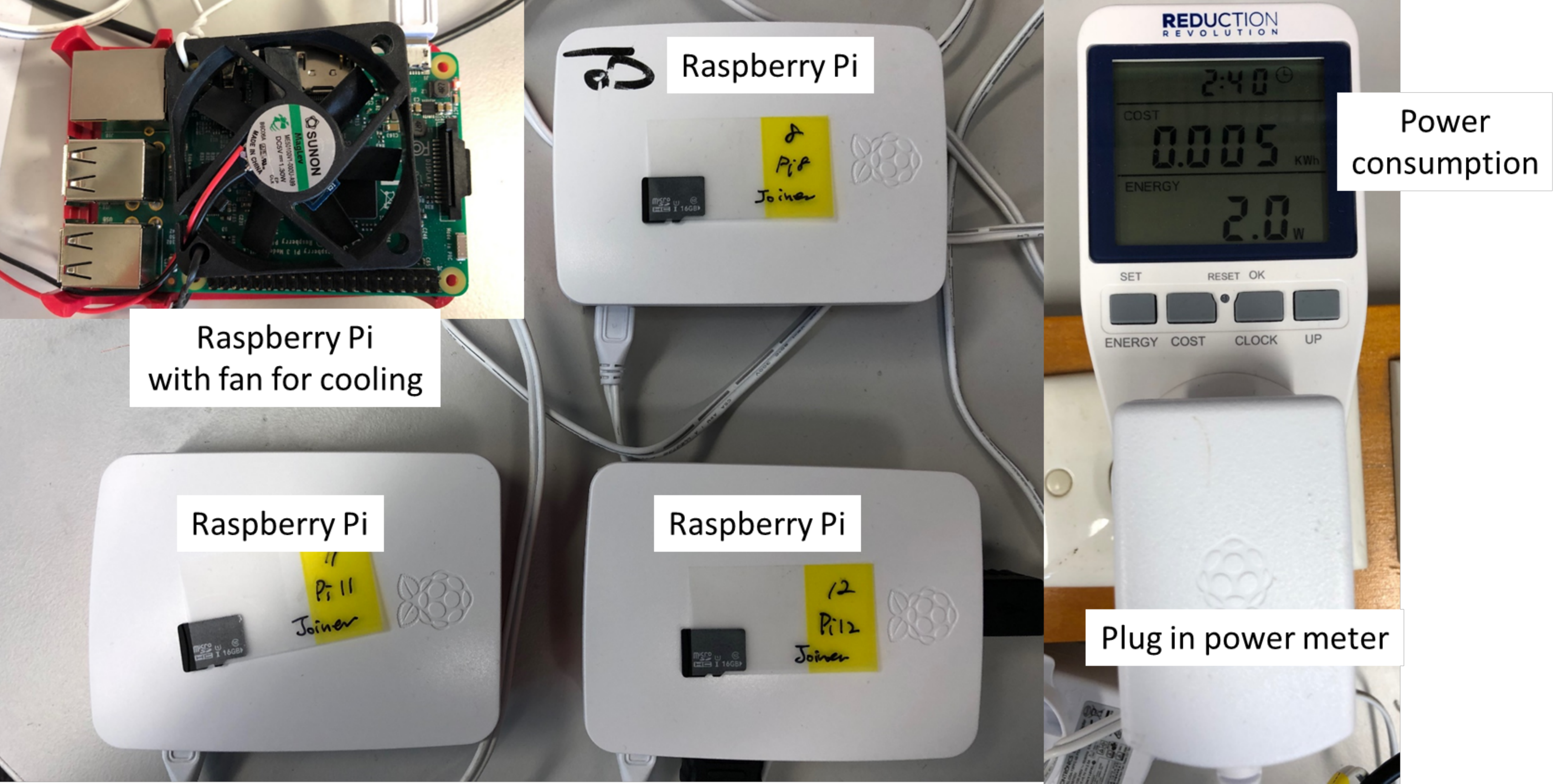}
	\caption{Four Raspberry Pi devices and a power meter are shown. 
% 	The top-right one shows how a small fan is attached for cooling. Plug-in powermeter measures the Raspberry Pi power consumption. 
	See the demo video for more details, \url{https://www.youtube.com/watch?v=x5mD1_EA2ps}.}
	\label{fig:setup}
\end{figure}

We use the Raspberry Pi 3 model BV1.2 (Fig.~\ref{fig:setup}) with the following settings: PyTorch version 1.0.0, OS Raspbian GNU/Linux 10 (buster), and Python version 3.7.3\footnote{We have made a unified manual guide of installing Pytorch v1.0.0 on Raspberry Pi at \url{https://github.com/Minki-Kim95/RaspberryPi}. We believe that this manual will help developers because we explain how to address the errors during installation, which are hard to resolve, and there are no solutions online.}. We note that CUDA is not available for the model. The server (laptop) has the following settings: CPU i7-7700HQ, GPU GTX 1050, Pytorch version 1.0.0, OS windows 10, Python version 3.6.8 using Anaconda, and the CUDA version 10.1.

\subsection{Measurement Methods of Performance Metrics}
% \garrison{Minki: which specific methods you used to measure the performance.}

\paragraph{Training Time} 

We use Python's \textsf{time} library to measure the training time containing the communication time between client and server. We set the time as $T_{\rm start}$ when the model starts training. Once the training is finished, we set the time as $T_{\rm end}$. As a result, training time is \mbox{$T_{\rm end} - T_{\rm start}$}. 
% Please note that the training time is influenced by the network traffic in practice, because both FL and SL require upstream and downstream communications.

\paragraph{Memory Usage}
We use Linux \texttt{free -h} command for measuring the memory usage. This command provides the memory information of \texttt{total}, \texttt{used}, \texttt{free}, \texttt{cached} and \texttt{available}. 
% Here the \texttt{available} is an estimation of available memory to start new applications, without swapping. It is roughly the sum of \texttt{free} and \texttt{cached}; \texttt{total} is the sum of \texttt{used} and \texttt{available}. 
The total memory of the Raspberry Pi device used in this experiment is 926~MB. The focus here is to record and report the \texttt{used} memory during training.

\paragraph{Power Consumption}
% There is no available python function to measure power consumption of Raspberry Pi. 

We use a plug-in powermeter, as shown in Fig.~\ref{fig:setup}, to measure the power consumption. We measured the power consumption in the kilowatt-hour (kWh) unit.   
% {\bf SURYA: I am not feeling right the way kwh is written; please check!}.

\paragraph{Temperature}

We use Python's \textsf{CPUTemperature}() function from the \texttt{CPUTemperature} library to monitor the temperature of the Raspberry Pi CPU. Notably, the device temperature could go high, e.g., 80$\celsius$ during training. Therefore, it may be necessary to cool down the device. A cooling fan can be attached to the Raspberry Pi (Fig.~\ref{fig:setup}). This practice can efficiently cool it down from 83$\celsius$ to 54$\celsius$.

\paragraph{Communication Overhead}
We measure the transmitted data size from each client to the server and vice versa. We use the \texttt{pickle} library to monitor the size of the transmitted data. We use the router DGN2200 v4 (N300 Wireless ADSL2+ Modem Router) for wireless communication between Raspberry Pi and the server.

% \subsubsection{Install Pytorch on Raspberry Pi}

% \subsubsection{Temperature of Raspberry Pi}

\subsection{Evaluation Considerations}
We considered the following three evaluation settings\footnote{We set \textit{one local epoch per round for FL and SFLV1} in all experiments. We compare implementation overhead by presetting a fixed number of \textit{100 rounds}. 
% In other words, 100 epochs for both of them. 
We always use the learning rate of 0.001.}:

\vspace{2pt}\noindent$\bullet$  We first Evaluate SL when a different number of layers is split and running on the client, given the same model architecture. Specifically, one, two, three layers are split and running on the client. This shows the advantage of SL to relax the device side computation overhead. (Section~\ref{sec:resultDifSplit}).

\vspace{2pt}\noindent$\bullet$ We then evaluate FL and SL across five clients with different model architectures. For the SL, two split layers run on clients regardless of model architectures. This shows the SL is invariant to the model complexity as long as the device sub-network is invariant. (Section~\ref{sec:resultDifModels}).

\vspace{2pt}\noindent$\bullet$ We further evaluate FL, SL, and SFLV1 across a range of clients from two to five with the same model architecture. This shows the SFLV1 on-device computation overheard inherits the advantage of SL, while greatly avoids its undesirable sequential training procedure among clients. (Section~\ref{sec:resultDifClient}).

For all evaluations, we report the performance overhead for a single Raspberry Pi device because we are interested in the client's overhead. 
\subsection{Effects of Number of Split Layers in SL}\label{sec:resultDifSplit}

Experiments are carried on five Raspberry Pi devices to observe the effect of the number of split layers for SL. All effects except the time overhead will apply to the SFLV1, as the latter by design should share the same \textit{computation} overhead with SL. 

\begin{figure*}[h]
	\centering
	\includegraphics[trim=0 0 0 0,clip,width=1.0\textwidth]{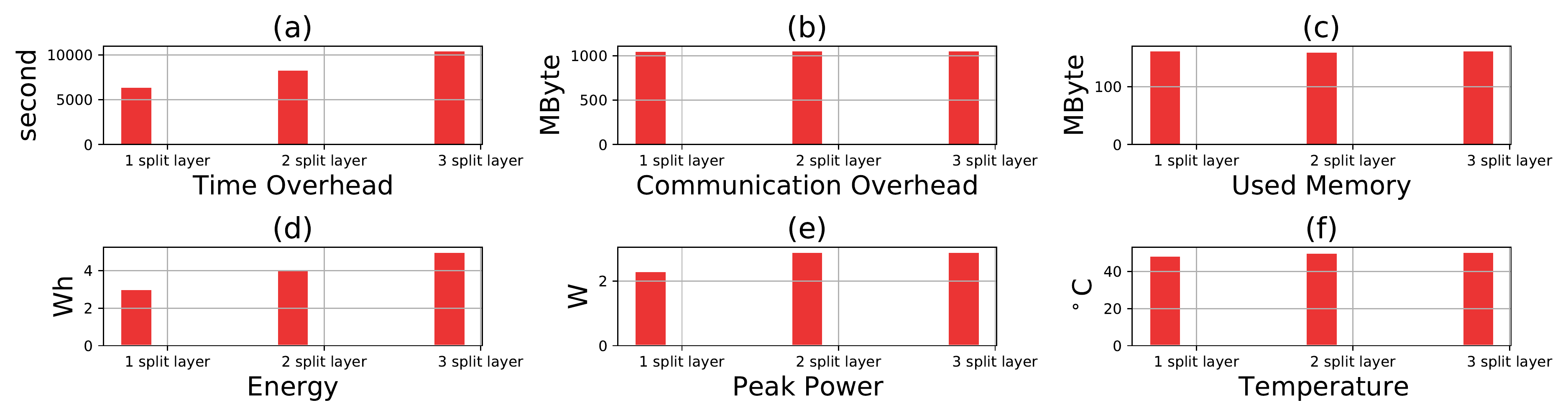}
	\caption{SL performance when a different number of split layers---1 to 3 convolutional layers---run on Raspberry Pi devices. All tests run {\it across 5 Raspberry Pi devices} and in a lab environment equipped with the 100 Gbit/s dedicated LAN. The model is with 4 convolutional layers and 2 dense layers.}
	\label{fig:PI_FLVsSL_differentLayers_ecg}
\end{figure*}

Results are detailed in Fig.~\ref{fig:PI_FLVsSL_differentLayers_ecg}. The communication overhead remains the same regardless of the number of layers at the client-side because the communication overhead in SL depends on the number of parameters in the cut layer rather than the number of split layers. The memory usage in Fig.~\ref{fig:PI_FLVsSL_differentLayers_ecg} (c), shows only a slight increase with the number of split layers. Most noticeably, time overhead (illustrated in  Fig.~\ref{fig:PI_FLVsSL_differentLayers_ecg} (a)) and energy overhead (depicted in Fig.~\ref{fig:PI_FLVsSL_differentLayers_ecg} (d)) increase with the number of split layers because the model complexity increases for training. Therefore, in practice, from the overhead reduction perspective, it is preferred to run a few layers only at the client-side for SL. 

\subsection{Effects of Model Complexity}\label{sec:resultDifModels}

To observe the effects of different models with varying complexity for SL and FL, we perform experiments on five Raspberry Pi devices. The models have a varying number of convolutional layers ranging from four to eight, thus varying model size. For SL, the on-device subnetwork is fixed.

\begin{figure*}[h]
	\centering
	\includegraphics[trim=0 0 0 0,clip,width=1.0\textwidth]{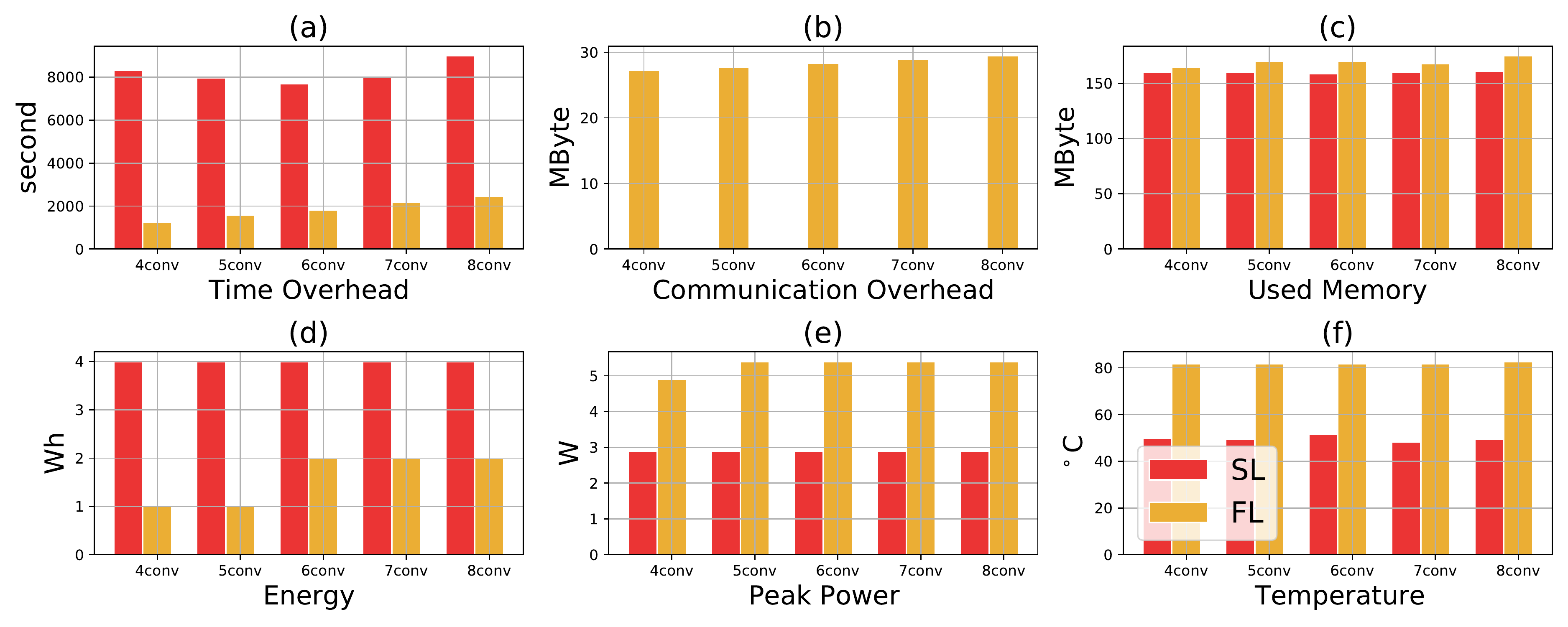}
	\caption{Overhead performance of FL and SL when the number of layers of the model varies from 4 conv to 8 conv, thus with differing model size. All tests were performed with the 100 Gbit/s dedicated LAN. For SL, the first two convolutional layers run at the client-side. For communication overhead, we only show the results for FL because SL's communication overhead (1054Mbytes) is not changed as well as significantly greater than FL's communication overhead, as shown in Fig.~\ref{fig:FLVsSL_Pis_ecg}.}
	\label{fig:PI_FLVsSL_differentModel_ecg}
\end{figure*}

According to the results depicted in Fig.~\ref{fig:PI_FLVsSL_differentModel_ecg}, the overhead, including time, communication, memory used, and energy consumed by Raspberry Pi devices (linearly), increases with the model complexity (defined by the number of layers in the model) for FL. In contrast, the overhead remains more or less constant for SL because the number of layers running on each client is fixed. Based on these findings, SL becomes more advantageous when we consider a model with higher complexity.

\subsection{Comparison among FL, SL, and SFLV1}\label{sec:resultDifClient}

\begin{figure*}[h]
	\centering
	\includegraphics[trim=0 0 0 0,clip,width=0.95\textwidth]{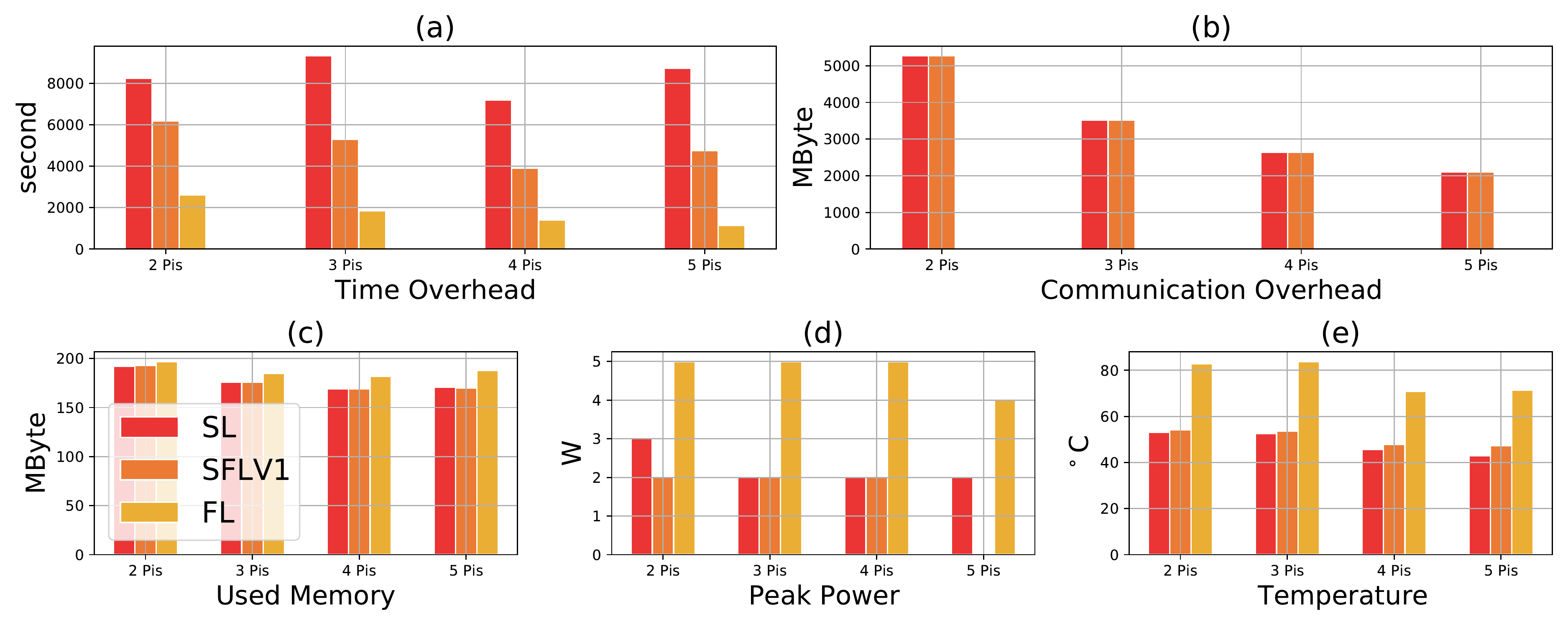}
	\caption{FL, SL and SFLV1 comparison when the number of Raspberry Pi devices (clients) varies from two to five.}
	\label{fig:FLVsSL_Pis_ecg}
\end{figure*}

Here, we evaluate and compare FL, SL, and SFLV1 when the number of Raspberry Pi devices (clients) ranges from two to five. Admittedly, the number of Pis implemented is not high. Therefore, this experiment (i) focuses on the overhead brought to individual Raspberry Pi rather than the server-side, and (ii) demonstrates the SFLV1 \textit{qualitative advantage} in terms of both computation reduction in comparison with FL and training time reduction in comparison with SL. One can easily extend the number of clients beyond five by adopting the released artifact (source code, user guide, and demo) of our experiment. The used model architecture has four 1D CNN layers and two dense layers. For SL as well SFLV1, the first two 1D CNN layers run on Raspberry Pi devices. It is worth mentioning that this implementation overhead is newly performed. The peak power under 2W cannot be measured by the powermeter (name: Bplug-S01, manufacturer: a-nine) used in this experiment; this explains the occasionally missing peak power of SLV1 in Fig.~\ref{fig:FLVsSL_Pis_ecg} (d).

The performance results are presented in Fig.~\ref{fig:FLVsSL_Pis_ecg}.  
% rather than the server perspective---assuming the server is always resource-rich.
As for the time overhead in Fig.~\ref{fig:FLVsSL_Pis_ecg} (a), FL reduces as the number of devices increases. This is due to the decrease in the local data size. SL slightly increases since each device runs the training sequentially. Overall, SL usually takes several times longer than the FL, given the same number of rounds. In terms of the SFLV1, we can see that its computation overheads, including communication, used memory, peak power, and temperature, are all the same as the SL, which are under expectation. Significantly, the training time is greatly reduced in comparison with the SL because now the IoT devices run in parallel. In general, this reduction becomes more obvious given that the number of devices is large. It is worth noting that the server in the experiment is eventually a personal laptop, and there is multiple-thread optimization made for the current SFLV1 evaluation. If the server is with a cluster of CPU and GPU cores, and the multiple-thread optimization is performed, we expect the training time will be close to the FL by fully exploiting the paralleling capability.

The communication overhead is presented in Fig.~\ref{fig:FLVsSL_Pis_ecg} (b)\footnote{We note that the communication overhead of FL is not displayed because it is orders of magnitudes smaller than that of SL, e.g., megabytes vs. gigabytes.}. FL stays relatively constant because the model parameters determine the FL's communication overhead rather than the local data size. SL and SFLV1 communication overhead decrease as it is highly related to the local data size. This corroborates with the statistical analysis result in a recent work~\cite{singh2019detailed}, where the communication overhead of SL is shown significantly higher than that of FL per round for low model complexity and fewer clients.

For the used memory, as shown in Fig.~\ref{fig:FLVsSL_Pis_ecg} (c), the FL is always higher than that of SL and SFLV1, because the FL needs to train the entire model in the Raspberry Pi, while the rest two only need to train a small subnetwork. This also leads to the high power peak, as shown in Fig.~\ref{fig:FLVsSL_Pis_ecg} (d), and high temperature of FL during training, as shown in Fig.~\ref{fig:FLVsSL_Pis_ecg} (e). In the FL. Without cooling, the Raspberry Pi device's temperature can be up to $83\celsius$ during the FL learning. 

However, although FL has a high power peak, the energy is lesser than that of SL for the same number of rounds, as shown in Fig.~\ref{fig:FLVsSL_Pis_ecg} (d). This is because, in FL, each client trains the local model {\it in parallel}, and consequently, the total time (accumulated computation and communication) for running a given number of rounds is less than that of SL.

\begin{mdframed}[backgroundcolor=black!10,rightline=false,leftline=false,topline=false,bottomline=false,roundcorner=2mm]
	\textbf{Notes:} (i) The computational overheads, including used memory, peak power, and temperature, of SL are always better than the FL. (ii) The training time and communication overhead of SL in our setting (small model, a limited number of participants) are worse than FL. (iii) SFLV1 shares the same overhead cost except for the training time, which is greatly reduced by the SFLV1 that is the main impetus of adopting in the IoT application.
\end{mdframed}

\section{Optimizations}\label{sec:optimization}
In this part, we demonstrate that the SFLG is an optimization of two specific SFL variants: SFLV1 and SFLV2. The SFLG allows the server to manage IoT devices when they are large-scaled flexibly. In addition, the communication overhead of the SFLG is optimized when the communication bandwidth is bottlenecked.
\subsection{SFLG Performance}\label{sec:SFvariant}
Though it is assumed that the server is always with rich-resources, it may still be undesirable to make a server-side sub-network copy corresponding to \textit{each device}, especially when the number of devices is becoming large, e.g., tens of thousands. Therefore, the scalability of SFLV1 could still be a concern for large-scale devices. In addition, it is shown that the learning performance of SFLV1 is close to FL, where it converges slowly in most cases  compared to SL. 

To improve the scalability and learning performance, we have proposed a generalized SFLG (as detailed in \autoref{sec:sflg}) that fits the IoT scenario with flexible configurations. Below, we demonstrate its advantage over two specific SFL variants of SFLV1 and SFLV2~\cite{splitfed}, which descriptions are referred to \autoref{sec:splitfed}.

% \subsubsection{SFLV2}
% \garrison{Chandra, see if you want to add the brief description of SFLV1 with its main motivation or advantage.}

% \subsubsection{SFLG}
% Now we detail the design of a new variant of SF, namely SFLG, which is a combination of SFLV1 and SFLV2 by utilizing the former's good parallelization and later's relaxed requirement of the number of server subnetwork.

As a recall, the implementation of SFLG is as below:
\begin{enumerate}
    \item the server makes a number of copies, $n$, of the server-side subnetwork, and each subnetwork copy is in charge of a group IoT devices.
    \item within a group of IoT devices, the SFLV2 is applied; whereas among the server-side subnetwork copies, the SFLV1 is used.
\end{enumerate}

In other words, within a group of IoT devices, one server-side subnetwork will sequentially make an update on each IoT device. On the other hand, different copies of the server-side network still run in parallel from the server's view.

\begin{figure}[t]
	\centering
	\includegraphics[trim=0 0 0 0,clip,width=0.5\textwidth]{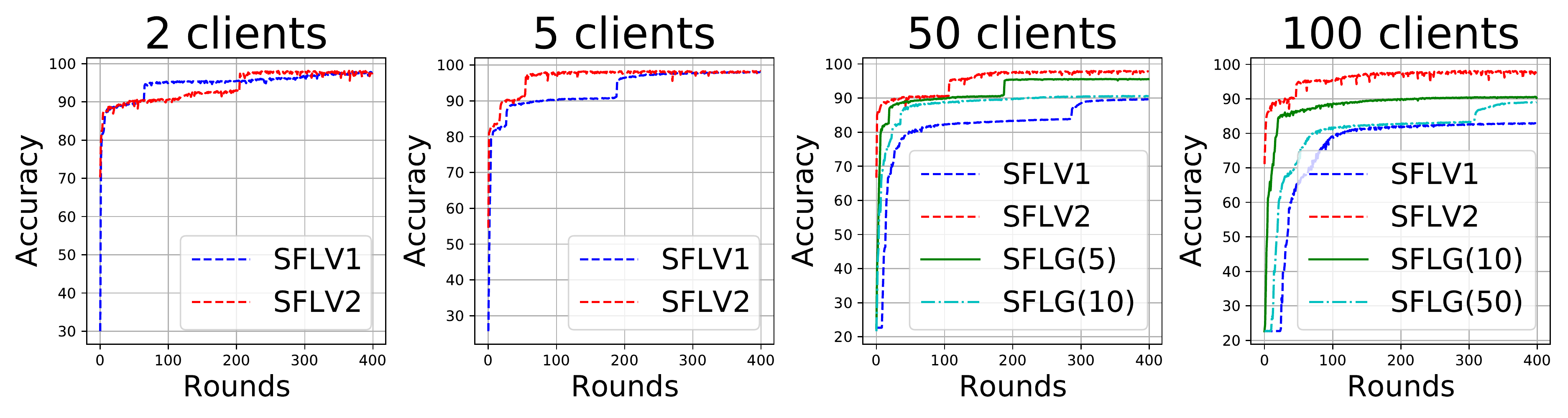}
	\caption{Testing accuracy SFLV1, SFLV2, and SFLG (\#) over rounds for the ECG data, which is IID and distributed in a balanced manner. \# indicates the number of copies made for the server-side subnetwork: smaller the \#, better scalability. In fact, (\# = 1) of the SFLG is SFLV2, while (\# = number of clients) of the SFLG is SFLV1.
	}
	\label{fig:SF_IID_multiple_clients}
\end{figure}

\begin{figure}[t]
	\centering
	\includegraphics[trim=0 0 0 0,clip,width=0.5\textwidth]{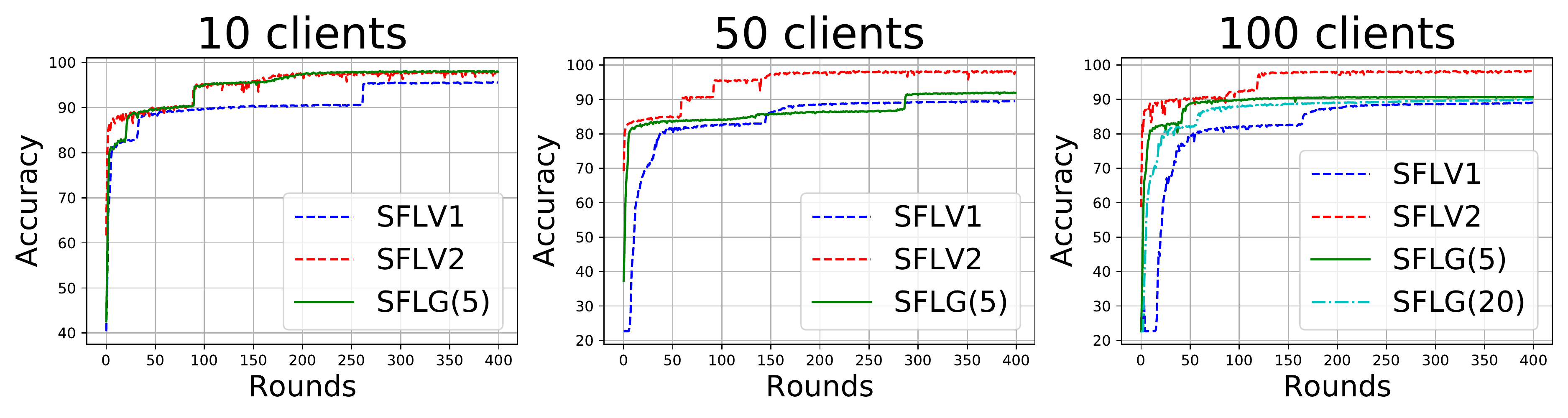}
	\caption{Testing accuracy SFLV1, SFLv2, and SFLG over rounds with imbalanced data setting on the ECG dataset.}
	\label{fig:SF_Imbalance}
\end{figure}

\begin{figure}[t]
	\centering
	\includegraphics[trim=0 0 0 0,clip,width=0.5\textwidth]{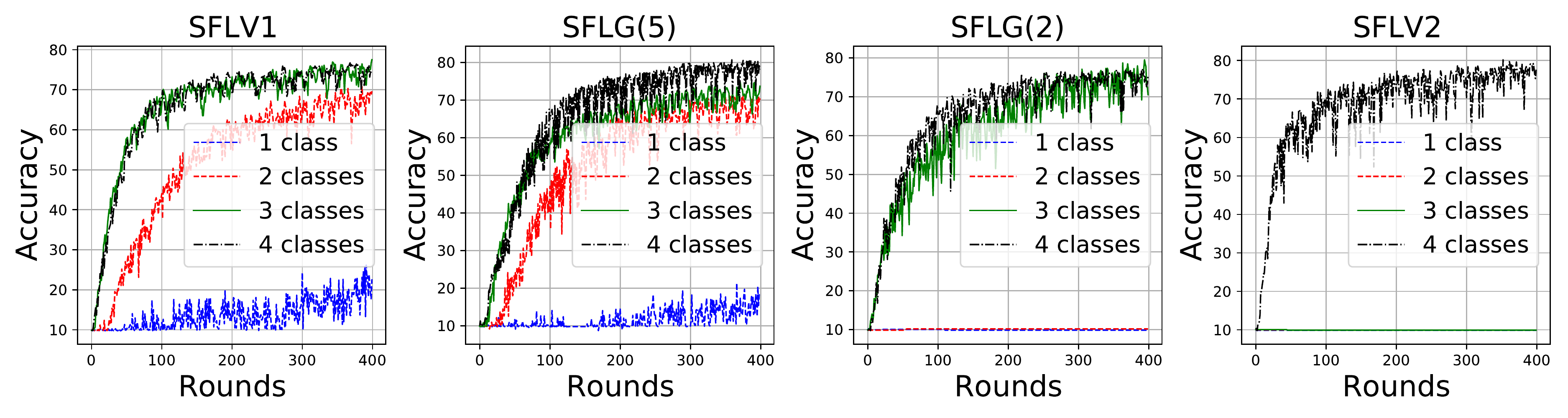}
	\caption{Testing accuracy of SFLV1, SFLV2, and SFLG over rounds with Non-IID dataset setting on the SC dataset. There are 10 clients in total, the SFLG(\#) means \# server-side subnetwork replicas are used. Each server-side subnetwork is in charge of $\frac{10}{\#}$ clients.} 
	\label{fig:SF_nonIID}
\end{figure}

\vspace{0.3cm}
\noindent{\bf Results:}
Now we compare the learning performance of split fed learning, including SFLG, the results are depicted in Fig.~\ref{fig:SF_IID_multiple_clients} with IID data distribution, Fig.~\ref{fig:SF_Imbalance} with imbalanced data distribution, and Fig.~\ref{fig:SF_nonIID} with non-IID data distribution.

For Fig.~\ref{fig:SF_IID_multiple_clients} with ECG dataset, we observe that the SFLV2 demonstrates the best learning performance as it converges the fastest, and the accuracy is also the highest given the same rounds used. In contrast, the SFLV1 is with the least performance. Interestingly, the SFLV2 in fact approaches the SL, while the SFLV1 approaches the FL. SFLV1 avoids the spikes that existed in the SL, demonstrating a more stable learning curve and no turn-down compared with Fig.~\ref{fig:IID_multiple_clients}. Similar trends are also true for Fig.~\ref{fig:SF_Imbalance} and Fig.~\ref{fig:SF_nonIID} under differing data distributions. In fact, when the number of the server-side network is set to be 1, the SFLG becomes SFLV2; when the number of the server-side network is equal to the number of IoT devices, the SFLG is SFLV1. This exactly corroborates our \textbf{Preposition 1} in \autoref{sec:sflg}. 
% Our key observation now is summarized as below.

\begin{mdframed}[backgroundcolor=black!10,rightline=false,leftline=false,topline=false,bottomline=false,roundcorner=2mm]
	\textbf{Notes:} The learning performance of the proposed SFLG is bounded by SFLV2 as upper-bound and SFLV1 as lower-bound given any tested data distribution settings. The SFLG provides flexibility in practice in terms of not only scalability but also learning performance under the IoT scenario. 
\end{mdframed}

As we have shown that the SFLV1 shares the learning performance with the FL (Fig.~\ref{fig:IID_multiple_clients_sc}), we have expected that the SFLV2 shares the learning performance with SL. While the latter is almost always the case, \textit{the SFLV2 is eventually with a relatively better performance with SL}. Since the SFLV2 (i) eliminates the unstable spikes of the learning curve exhibited in the SL and (ii) avoids the turn-down learning curve of the SL. The potential reason lies in the fact that the device-side subnetworks now have an average operation, which enhances the feature representations sharing learned across clients.

In practice, when the IoT devices are large-scaled, e.g., millions, the $|G|$ (number of server-side subnetwork copies) needs to be carefully chosen. Let $M$ be the memory required for 1 subnetwork copy at the server-side, now if there are $|G| \ge 1$ models, then there is $|G|\times M$ memory required to store the copies. Without SFLG that enables flexible chosen of |G|, the intuitive setting of making a server-side subnetwork copy is still prohibitive for the server even considering its rich-resource when the IoT devices are large-scaled. With SFLG, an acceptable $|G|$ can be chosen to ensure that large-scale IoT devices can be handled while still providing reasonable parallelization.

% For k large and large model kM is very large to fit within the server-side resources.

\subsection{SL and SFLG Communication Overhead Reduction}
Usually in practice (e.g., IoT environment), communication is a bottleneck. So it is imperative to reduce the communication in the IoT settings for the SL. Notably, \textit{all the communication overhead optimization directly applies to approaches leveraging split architectures, including SFLG}. We are interested in the following question.

\vspace{1mm}
\noindent{\it Is there any efficient means of reducing SL communication overhead while retaining the accuracy?}
\vspace{1mm}

The investigations below are affirmative by reducing the intermediate model output (cut layer) data size required to be communicated between the device and the server.

Given an input size of $A_{\rm in}$, the output size after the pooling layer is expressed:
\begin{equation}
    A_{\rm out} = \frac{A_{\rm in} - f}{s} + 1
\end{equation}
with $f$ the kernel size and the $s$ the stride of the pooling layer.

In general, the communication overhead is linear to the ratio of:
\begin{equation}\label{eq:factor}
    \text{factor} = \frac{A_{\rm out}}{A_{\rm in}}
\end{equation}
with factor measures the communication overhead reduction. Smaller it, better or less communication overhead. Therefore, given the $A_{\rm in}$, it is preferable to have a small $A_{\rm out}$ by tuning the pooling layer parameter to facilitate communication.

The key is to reduce the output size of the cut layer on the client-side. Both the convolutional layer and pooling layers can affect the output size given input size by using different parameter settings, e.g., kernel size $f$ and stride $s$. We can flexibly select suitable parameter settings that reduce the cut layer's output size while retaining the model accuracy. Below, we focus on the pooling layer for extensive experimental validations. We consider three settings: 

\vspace{2pt}\noindent$\bullet$ {\bf Setting 1:} Rearrange or move the pooling layers of the server subnetwork to the device.

\vspace{2pt}\noindent$\bullet$ {\bf Setting 2:} Tuning the pooling parameter in the device subnetwork while fixing the server subnetwork.

\vspace{2pt}\noindent$\bullet$ {\bf Setting 3:} Fixing the device subnetwork while tunning server subnetwork to improve accuracy if Setting 1 and 2 incur accuracy drop.

\begin{table*}
	\centering 
	\caption{Tuning the split model architecture to reduce communication overhead.}
    \resizebox{0.80\textwidth}{!}{
	\begin{tabular}{c | c | c | c | c | c} % 
		\toprule % Top horizontal line
		\toprule % Top horizontal line
		No. & Device Network & Server Network & \begin{tabular}{@{}c@{}} Comm. \\ Overhead \\ (Reduced by) \end{tabular} & factor & Accuracy \\ % Column names row
		 
		\midrule
        \begin{tabular}{@{}c@{}} 1 \\ (Baseline) \end{tabular} & \begin{tabular}{@{}c@{}} conv1d(1,16,7) \\ + Pool(2,2) \\ + conv1d(16,16,5) \end{tabular} & \begin{tabular}{@{}c@{}} conv1d(16,16,5) \\ + conv1d(16,16,5) \\ + Pool(2,2) \\+ Linear(25$\times$16, 128) + Relu \\ + Linear(128, 5) + Softmax \end{tabular} & 19,770 MB & N/A & 96.51\% \\
        \hline 

        \begin{tabular}{@{}c@{}} 2 \\ (Setting 1) \end{tabular} & \begin{tabular}{@{}c@{}} conv1d(1,16,7) \\ + Pool(2,2) \\ + conv1d(16,16,5) \\ + \textbf{Pool(2,2)} \end{tabular} & \begin{tabular}{@{}c@{}} conv1d(16,16,5) \\ + conv1d(16,16,5) \\ + \st{\textsf{Pool(2,2)}} \\+ Linear(21$\times$16, 128) + Relu \\ + Linear(128, 5) + Softmax \end{tabular} & \begin{tabular}{@{}c@{}} 9,938 MB \\ (50.26\%) \end{tabular} & \begin{tabular}{@{}c@{}} 50\% \\ ($\frac{29}{58}$) \end{tabular}  & 95.40\% \\ \hline

        \begin{tabular}{@{}c@{}} 3 \\ (Setting 3) \end{tabular} & \begin{tabular}{@{}c@{}} conv1d(1,16,7) \\ + Pool(2,2) \\ + conv1d(16,16,5) \\ + \textbf{Pool(2,2)} \end{tabular} & \begin{tabular}{@{}c@{}} conv1d(16,\textbf{32},5) \\ + conv1d(\textbf{32},\textbf{32},5) \\+ Linear(\textbf{21}$\times$\textbf{32}, 128) + Relu \\ + Linear(128, 5) + Softmax \end{tabular} & \begin{tabular}{@{}c@{}} 9,938 MB \\ (50.26\%) \end{tabular} & \begin{tabular}{@{}c@{}} 50\% \\ ($\frac{29}{58}$) \end{tabular} & 97.91\% \\ \hline
        
        \begin{tabular}{@{}c@{}} 4 \\ (Setting 2) \end{tabular} & \begin{tabular}{@{}c@{}} conv1d(1,16,7) \\ + Pool(2,2) \\ + conv1d(16,16,5) \\ + \textbf{Pool(4,2)} \end{tabular} & \begin{tabular}{@{}c@{}} conv1d(16,\textbf{32},5) \\ + conv1d(\textbf{32},\textbf{32},5) \\+ Linear(\textbf{20}$\times$\textbf{32}, 128) + Relu \\ + Linear(128, 5) + Softmax \end{tabular} & \begin{tabular}{@{}c@{}} 9,599 MB \\ (48.55\%) \end{tabular} & \begin{tabular}{@{}c@{}} 48.28\% \\ ($\frac{28}{58}$) \end{tabular} & 96.63\% \\  \hline
        
        \begin{tabular}{@{}c@{}} 5 \\ (Setting 2) \end{tabular} & \begin{tabular}{@{}c@{}} conv1d(1,16,7) \\ + Pool(2,2) \\ + conv1d(16,16,5) \\ + \textbf{Pool(6,2)} \end{tabular} & \begin{tabular}{@{}c@{}} conv1d(16,\textbf{32},5) \\ + conv1d(\textbf{32},\textbf{32},5) \\+ Linear(\textbf{19}$\times$\textbf{32}, 128) + Relu \\ + Linear(128, 5) + Softmax \end{tabular} & \begin{tabular}{@{}c@{}} 9,260 MB \\ (46.84\%) \end{tabular} & \begin{tabular}{@{}c@{}} 46.55\% \\ ($\frac{27}{58}$) \end{tabular} & 96.96\% \\ \hline      

        \begin{tabular}{@{}c@{}} 6 \\ (Setting 2) \end{tabular} & \begin{tabular}{@{}c@{}} conv1d(1,16,7) \\ + Pool(2,2) \\ + conv1d(16,16,5) \\ + \textbf{Pool(8,2)} \end{tabular} & \begin{tabular}{@{}c@{}} conv1d(16,\textbf{32},5) \\ + conv1d(\textbf{32},\textbf{32},5) \\+ Linear(\textbf{18}$\times$\textbf{32}, 128) + Relu \\ + Linear(128, 5) + Softmax \end{tabular} & \begin{tabular}{@{}c@{}} 8,951 MB \\ (45.27\%) \end{tabular} & \begin{tabular}{@{}c@{}} 44.82\% \\ ($\frac{27}{58}$) \end{tabular} & 95.61\% \\ \hline      

        \begin{tabular}{@{}c@{}} 7 \\ (Setting 2) \end{tabular} & \begin{tabular}{@{}c@{}} conv1d(1,16,7) \\ + Pool(2,2) \\ + conv1d(16,16,5) \\ + \textbf{Pool(2,4)} \end{tabular} & \begin{tabular}{@{}c@{}} conv1d(16,\textbf{32},5) \\ + conv1d(\textbf{32},\textbf{32},5) \\+ Linear(\textbf{7}$\times$\textbf{32}, 128) + Relu \\ + Linear(128, 5) + Softmax \end{tabular} & \begin{tabular}{@{}c@{}} 5,192 MB \\ (26,26\%) \end{tabular} & \begin{tabular}{@{}c@{}} 25.86\% \\ ($\frac{15}{58}$) \end{tabular} & 95.23\% \\ \hline
        
        \begin{tabular}{@{}c@{}} 8 \\ (Setting 3) \end{tabular} & \begin{tabular}{@{}c@{}} conv1d(1,16,7) \\ + Pool(2,2) \\ + conv1d(16,16,5) \\ + \textbf{Pool(2,4)} \end{tabular} & \begin{tabular}{@{}c@{}} conv1d(16,\textbf{32},5) \\ + conv1d(\textbf{32},\textbf{64},5) \\+ Linear(\textbf{7}$\times$\textbf{64}, 128) + Relu \\ + Linear(128, 5) + Softmax \end{tabular} & \begin{tabular}{@{}c@{}} 5,192 MB \\ (26,26\%) \end{tabular} & \begin{tabular}{@{}c@{}} 25.86\% \\ ($\frac{15}{58}$) \end{tabular} & 97.54\% \\   
        
        % 2 & \begin{tabular}{@{}c@{}} conv1d(1,16,7) \\ + Pool(2,2) \\ + conv1d(16,16,5) \\ + \textbf{Pool(2,4)} \end{tabular} & \begin{tabular}{@{}c@{}} conv1d(16,\textbf{32},5, \textbf{2}) \\ + conv1d(\textbf{32},\textbf{64},5, \textbf{2}) \\+ Linear(\textbf{15}$\times$\textbf{64}, 128) + Relu \\ + Linear(128, 5) + Softmax \end{tabular} & \begin{tabular}{@{}c@{}} 5,192 MB \\ (26,26\%) \end{tabular} & 25.86\% & 97.66\% \\                    
        \hline 
        
		\bottomrule
	\end{tabular}
	}
	\label{tab:Setting2} 	 \begin{tablenotes}
      \small
      \item 1. The leakyrelu activation is used after \textit{each conv1d layer}, but not shown for concise purpose.
      \item 2. Device communication (Comm.) overhead is the data size sent to and received from the server. The number of epochs is 200.
      \item 3. conv1d(\#input channel, \#output channel, \#kernel size, \#padding). Pool(\#kernel size, \#stride). The output size of the second conv1d is $58\times 16$ given the input size to the network is $1\times 130$.
    \end{tablenotes}
\end{table*}

\subsubsection{Setting 1} In this setting, we consider moving the pooling layer in the server subnetwork to the device side. In particular, we move it to the cut-layer of the device side, which is detailed in Tabel~\ref{tab:Setting2} (No.2 compared with No.1 as baseline). We can see $2\times$ reduction in communication overhead. Note that, due to the position change of the pooling layer, the input size to the first dense layer (or fully connected layer) at the server subnetwork is also changed. As for the accuracy, we observe a slight decrease of about 1\%, which tends to be acceptable considering the significant communication overhead decrease. If it does matter, this slight accuracy drop can be compensated via setting 3 detailed soon.

\subsubsection{Setting 2} In this setting, we tune the pooling layer parameter as detailed in Tabel~\ref{tab:Setting2}. As shown in No.4, 5, and 6, the kernel size of the pooling is changed from 2 to 4, 6, and 8, respectively, while retaining the stride being unchanged. This gradually decreases the communication overhead to 48.28\%, 46.55\%, and 44.82\%. 

In No.7, the stride is changed from 2 to 4 while kernel size is retained to be 2. Such stride change results in significant communication overhead reduction---$4\times$ reduction in comparison with baseline in No.1. Notably, the server subnetwork does require to change the input size of the first dense layer for the sake of size compatibility. As for the accuracy, it slightly drops. Again, this can be compensated by below setting 3.

\subsubsection{Setting 3} This is to compensate for the slight accuracy drop resulted from setting 1 or/and 2 if the accuracy requirement is stringent. The insight here is to only fine-tune the model architecture in the server subnetwork without bringing any overhead to the device side. As detailed in No.3 and 8, the number of filters is increased in the two conv1d layers---making the server subnetwork to be wider. Consequentially, the accuracy is improved, which is in fact slightly higher than the baseline in No.1. In this context, the communication overhead reduction is reserved while the accuracy is retained without any additional overhead brought to the device.
% We considered and tried two means for achieving so. The first is to use dropout layer in the last layer of the client subnetwork. However, we found that the dropout hurts the model accuracy since the model here is small, while dropout is usually used a a regularization technique to mitigate overfitting for large model.

In Tabel~\ref{tab:Setting2}, we have also experimentally measured the communication overhead besides the theory analysis in \eqref{eq:factor}. While they agree well, the experimental reduction is slightly higher than the analysis in \eqref{eq:factor}. The reason is that the subnetwork on the device-side needs to be sent to the server-side for updating per epoch, which results in additional small communication overhead that is not counted by the analysis in \eqref{eq:factor}. Overall, by properly configuring the network resided in the device-side (e.g., pooling layers) and in the server-side (e.g., network width), we have experimentally validated that the communication overhead in SL can be substantially decreased without affecting the model accuracy and bringing a burden to the resource-restrict device. As the training time and power consumption are also linearly related to the communication overhead given the same communication bandwidth. Reducing the communication overhead will also reduce the training time and power consumption.

\begin{mdframed}[backgroundcolor=black!10,rightline=false,leftline=false,topline=false,bottomline=false,roundcorner=2mm]
	\textbf{Notes:} Unlike the FL which communication overhead is agnostic to the localized data---it is only dependent on the given (entire) model size, SL and SFLG is dependent on the size of the cut layer and the number of forward and backward propagation, where the later is related to the size of the localized data. 
\end{mdframed}

Therefore, it is worth to consider i) means of reducing the size of the cut layer as we have validated and ii) further reduce the size of the localized data in future investigations. The latter can be realized by, e.g., principal component analysis, to reduce the feature size to the data before feeding to the local subnetwork. This could have two benefits: besides communication overhead reduction, reducing the on-device computation over input with low feature dimensions.

\textit{Note for the SFLG, it can exploit the edge-computing resource such that an edge server manages a group of neighboring IoT devices (similar to SFLV2) to relax the communication bandwidth, while these edging servers run parallel for training time expedition (similar to SFLV1).}  

In addition, it is expected that the 5G with the high data rate, more bandwidth will be widely used in future---it is readily becoming available~\cite{chettri2019comprehensive}. In this context, it is desirable to choose SFLG for large-scaled IoT applications as it relaxes the on-device computation overhead. 

\section{Conclusion}
\label{sec:Conclusion}

This work is the first to empirically evaluate, and end-to-end compare FL, SL, and SFL in real-world IoT settings. We comprehensively assessed the learning performance in terms of model accuracy and convergence speed of each of them. 
We mainly considered imbalanced and non-IID distributions for our experiments, which would be more realistically resemble IoT scenarios. The FL and SL have their own benefits and shortcomings in terms of learning performance. Our devised SFLG takes advantage of SL and FL to better fit IoT applications.
% , in particular, reducing the on-device computational overhead while expediting training time.
% Similar to FL, SL is also inevitably influenced by data distribution. In general, SL performs better than FL in the case of imbalanced data distributions but can rather worsen than FL in the case of extreme non-IID data distributions.
Besides, we extensively evaluated the practicality of mounting the \textit{training} of FL, SL, and SFL on resource-restricted  computing platforms, Raspberry Pis. We mainly dealt with pervasive sequential time-series data and provided useful comprehensive results---various implementation overhead---to the community. We corroborate the design impetus of SFLG to provide (i) better computational overhead compared than FL and (ii) greatly training time reduction compared with SL. We have further validated the flexibility of the proposed SFLG in comparison with two specific SFL variants. By considering the possible communication bottleneck in practice (e.g., IoT settings), we proposed pragmatic approaches to reduce the communication overhead of SL (also applicable for SFLG) and empirically validated the efficiency of those optimizations. 
% Overall, for the IoT scenario, the FL would be a more practical recommendation because it requires less overall communication, time, and power consumption overhead when a simple 1D CNN model is used. However, we also found that for both FL and SL, the use of more complicated models would still be infeasible to mount {\it training} on low-capacity IoT devices such as Raspberry Pi.

%\section*{Acknowledgment}
%%\begin{small}
%The work has been supported by the Cyber Security Research Centre Limited whose activities are partially funded by the Australian Government’s Cooperative Research Centres Programme. 
% This work was also supported by NRFK (2019R1C1C1007118).
% This work was also supported in part by the ITRC support program (IITP-2019-2015-0-00403). The authors would like to thank all the anonymous reviewers for their valuable feedback.
%\end{small}
%
\balance
\end{document}